\documentclass[manuscript,screen]{acmart}

\usepackage{amsmath, amsfonts}
\usepackage{graphicx}
\usepackage{colortbl}

\usepackage{longtable}
\usepackage{afterpage}

\usepackage{multirow}
\usepackage{adjustbox}
\usepackage{hyperref}
\usepackage{tikz}
\usepackage{graphicx}
\usepackage[utf8]{inputenc}
\usepackage[T1]{fontenc}
\usepackage{cleveref}
\usepackage{caption}
\usepackage{subcaption}
\usepackage{footnote}
\usepackage{tablefootnote}

\usepackage[none]{hyphenat}

\usepackage{pifont}
\usepackage{xfrac}

\usepackage{svg}
\usepackage{xspace}

\usepackage[inline]{enumitem}

\newcommand{\attackMitigatedMark}{\text{\ding{51}}}%
\newcommand{\defenceBrokenMark}{\text{\ding{55}}}%
\newcommand{\attackLoweredMark}{$\searrow$\xspace}%
\newcommand{\speculatemark}{$\wr$}%
\newcommand{\attackClaimMark}{$\thicksim$}%
\newcommand{\nonap}{\cellcolor{lightgray}}%
\newcommand{\unknown}{N/A}%

\newcommand{\fully}{\checkmark}%
\newcommand{\partially}{$\thicksim$}%

\usepackage{array}

\newcolumntype{C}[1]{>{\centering\let\newline\\\arraybackslash}m{#1}}

\crefformat{footnote}{#2\footnotemark[#1]#3} 

\setcopyright{acmlicensed}
\acmJournal{CSUR}
\acmYear{2023} \acmVolume{1} \acmNumber{1} \acmArticle{1} \acmMonth{1} \acmPrice{15.00}\acmDOI{10.1145/3595292}

\begin{document}

%

\title{I Know What You Trained Last Summer:
A Survey on Stealing Machine Learning Models and Defences}


\author{Daryna Oliynyk}
\affiliation{%
  \institution{SBA Research}
  \city{Vienna}
  \country{Austria}}
\email{doliynyk@sba-research.org}

\author{Rudolf Mayer}
\affiliation{%
  \institution{SBA Research \& Vienna University of Technology}
  \city{Vienna}
  \country{Austria}}
 \email{rmayer@sba-research.org}

\author{Andreas Rauber}
\affiliation{%
  \institution{Vienna University of Technology}
  \city{Vienna}
  \country{Austria}}
 \email{rauber@ifs.tuwien.ac.at}

\begin{abstract}
  
Machine-Learning-as-a-Service (MLaaS) has become a widespread paradigm, making even the most complex Machine Learning models available for clients via e.g. a pay-per-query principle. This allows users to avoid time-consuming processes of data collection, hyperparameter tuning, and model training. However, by giving their customers access to the (predictions of their) models, MLaaS providers endanger their intellectual property such as sensitive training data, optimised hyperparameters, or learned model parameters.
In some cases, adversaries can create a copy of the model with (almost) identical behaviour using the the prediction labels only. While many variants of this attack have been described, only scattered defence strategies that address isolated threats have been proposed.
To arrive at a comprehensive understanding why these attacks are successful and how they could be holistically defended against, a thorough systematisation of the field of model stealing is necessary.
We address this by categorising and comparing model stealing attacks, assessing their performance, and exploring corresponding defence techniques in different settings. We propose a taxonomy for attack and defence approaches and provide guidelines on how to select the right attack- or defence strategy based on the goal and available resources.
Finally, we analyse which defences are rendered less effective by current attack strategies.

\end{abstract}



\maketitle

\section{Introduction}

Training a Machine Learning model can be very complex and time- as well as resource-consuming. To safeguard their intellectual property, owners may opt to keep their models secret, allowing external users to access them only by input-output queries over a predefined API. However, black-box access to a model does not imply a \textit{protected} model. Recent work has shown how an adversary can \textit{steal} (extract) such models \cite{tramer_stealing_2016,papernot_practical_2017,orekondy_knockoff_2019}.
The technique of model stealing (also called "model extraction") aims at obtaining e.g. training hyperparameters, the model architecture, learned parameters, or an approximation of the behaviour of a model, all of which to the detriment of the lawful model owner.

The number of domains where model stealing attacks are successful has dramatically risen over the last few years. Dozens of attacks were executed regarding attack image classification \cite{orekondy_knockoff_2019}, text classification \cite{yi_shi_how_2017}, natural language processing \cite{krishna_thieves_2020}, and reinforcement learning \cite{behzadan_adversarial_2019}. Jagielski et al. provide a preliminary taxonomy based on the attackers' goals, thus classifying different types of attacks \cite{jagielski_high_2020}. However, the authors focus on a specific subset of attack patterns that address behaviour stealing and target only neural networks. Therefore, a comprehensive analysis of the potential and abilities of model stealing remains an important open task. 

There are two main approaches for protecting a Machine Learning model against a model stealing attack: attack detection \cite{juuti_prada_2019} and attack prevention \cite{orekondy_prediction_2020}. The first approach cannot protect the model on its own, but informs the owner that somebody tries to steal the model or that it has already been stolen. The second approach should prevent the attack or at least make it less effective. Unfortunately, a lot of the defences can either be fooled \cite{correia-silva_copycat_2018,atli_extraction_2020,mosafi_stealing_2019} or work only under specific conditions \cite{yan_monitoring-based_2021}. Hence, studying existing approaches and investigating new ones is of the utmost importance. To the best of our knowledge, there is no work that comprehensively compares defences against model stealing. A systematisation of defence approaches will lead to a better understanding of success criteria and, subsequently, to new, more effective defences -- for instance, by combining defences to cover multiple attack models at once.

Our contributions in this paper are the following: 
\begin{itemize}
    \item We collect and describe approaches to model stealing attacks as well as defence techniques. We explore how, when, and for which goals they were created, and unify reported performance measures of known attacks.
    \item We provide novel a taxonomy of model stealing attacks and -defences based on goal, methodology, and target model type. Following this taxonomy, we classify attacks and defences.
    \item We compare \textit{query-based} attacks by their effectiveness and efficiency; based on this comparison we develop recommendations for how to design and evaluate model stealing attacks. 
    \item We provide two guidelines for model stealing attacks and -defences, illustrated with diagrams. Following those guidelines, one can decide which attack- or defence strategy suits best in a given setting.
\end{itemize}

The rest of the paper is structured as follows:
\Cref{sec:related_surveys} describes related work, before \Cref{sec:methodology} details the methodology used in our survey and systematisation.
\Cref{sec:background} provides the reader with the background knowledge required to understand this paper, while \Cref{sec:modelStealing} introduces important concepts of model stealing.
\Cref{sec:model_stealing} introduces our novel taxonomy on model stealing attacks, followed by
\Cref{sec:query-based,sec:side-channel} which describes known attack approaches and provides the corresponding classification of the attacks as well as an overview of the performance of individual attacks. \Cref{sec:defences} describes proposed defence strategies and provides a respective taxonomy; subsequently, \Cref{sec:defences-vs-attacks} presents two guidelines for choosing the best attack- or defence strategy under certain conditions and compares the effectiveness of defences against known attacks.
Finally, \Cref{sec:conclusion} provides conclusions and an outlook for future work.

\section{Related Work}
\label{sec:related_surveys}
To the best of our knowledge, there is no systematisation that provides a comprehensive research of model stealing attacks and defence techniques. Jagielski et al. \cite{jagielski_high_2020} were the first to categorise model stealing attacks in terms of two objectives: accuracy and fidelity. The authors compared the goals of different attacks and argued about the importance of fidelity, which is a valuable basis for this work. However, they focus only on a specific subset of attacks, i.e. behaviour stealing of neural networks, and do not include defence strategies. In our work, we propose a comprehensive taxonomy and systematisation of both attacks and defences. Given the dynamics of the field, we are also able to consider a significantly larger number of papers (more than $100$ on attacks and on defences, as opposed to $9$ on attacks). 

Gong et al.~\cite{gong_model_2020} provide an overview of six model stealing attacks as well as six defences. The authors categorised them based on specific characteristics, e.g. an ability to steal/protect a deep neural network (DNN). However, the paper covers only a fraction of relevant works, and consequently the taxonomy and categorisation comprises only a subset of the field.
In contrast, a significant number of surveys regarding privacy and security in Machine Learning have been published \cite{papernot_sok_2018,qiu_review_2019,shafique_robust_2020, he_towards_2021}. In these works, model extraction attacks are usually only briefly mentioned as one sub-field of adversarial Machine Learning, while the main focus is on e.g. evasion (adversarial examples) and data poisoning attacks. Furthermore, there are publications which explore attacks and defences, including model stealing, in specific settings like reinforcement learning \cite{ilahi_challenges_2021} or edge-deployed neural networks \cite{isakov_survey_2019}.
We go beyond these studies and focus on model stealing as a crucial issue of Machine Learning security, presenting a comprehensive, structural view on the broad range of attacks as well as defences.

\section{Methodology}\label{sec:methodology}
Our paper is based on an extensive literature research, including \textit{formal}, peer-reviewed literature such as conference papers or journal articles as well as \textit{grey} literature, i.e. works that did not undergo a peer-review process; the latter primarily includes pre-prints published on the arXiv repository.

We defined the following criteria to identify the most relevant literature regarding model stealing. Our inclusion criteria are:
\begin{enumerate*}[label=(\arabic*)]
    \item Literature which proposes a method to perform model stealing attacks
    \item Literature which proposes a defence against model stealing attacks.
    \item Literature which evaluates or compares earlier schemes.
\end{enumerate*}

Our exclusion criteria are:
\begin{enumerate*}[label=(\arabic*)]
    \item (Near) Duplicates; if the titles are different, but the content is very similar, we consider the most comprehensive or peer-reviewed version, and cite only that version.
    \item Literature which only \textit{applies} earlier model stealing attacks as vehicle, without introducing novel attacks or -defences. This includes e.g. using model stealing to transform black-box access to a model into white-box access to a substitute model for an evasion attack.
\end{enumerate*}

This resulted in a total of more than $100$ papers on model stealing attacks and -defences for our in-depth investigation.
\begin{figure}
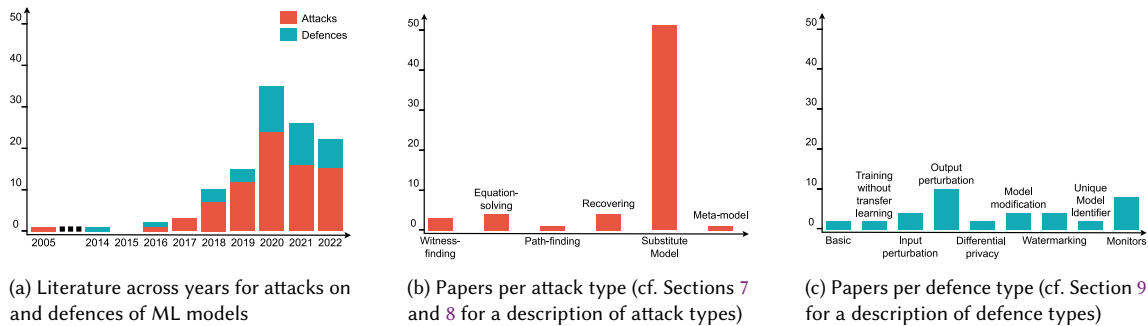

     \centering
     \begin{subfigure}{0.3\textwidth}
         \centering
         \includesvg[width=\textwidth]{figures-literature/literature_by_years.svg}
 \caption{Literature across years for attacks on and defences of ML models}
 \label{fig:distribution-years-attack-defence}
     \end{subfigure}
     \hfill
     \begin{subfigure}{0.3\textwidth}
         \centering
         \includesvg[width=\textwidth]{figures-literature/literature_attacks.svg}
         \caption{Papers per attack type (cf. \Cref{sec:query-based,sec:side-channel} for a description of attack types)}
         \label{fig:papers-per-attack-type}
     \end{subfigure}
     \hfill
     \begin{subfigure}{0.3\textwidth}
         \centering
         \includesvg[width=\textwidth]{figures-literature/literature_defences.svg}
         \caption{Papers per defence type (cf. \Cref{sec:defences} for a description of defence types)}
         \label{fig:papers-per-defence-type}
     \end{subfigure}
     \caption{Literature statistics
     }
\end{figure}
\Cref{fig:distribution-years-attack-defence} provides an overview on how the field evolved over the years. There is very early work from 2005 (\cite{lowd_adversarial_2005}) which covers model stealing, but the main body of literature has been published since 2016. First, more work was published on attacks; however, the volume of literature on defence strategies has caught up since 2018. It has to be noted that some publications cover an attack and the respective defence as well as propose a new defence immune to that attack.
Regarding the different types of attacks depicted in \Cref{fig:papers-per-attack-type}, we can see that the vast majority of works focuses on substitute model training attacks. Since there are many attacks utilising hardware (HW) or software (SW) side channels (SCA), it must be noted that these attacks often exploit multiple, different types of side-channels, thus forming a rather inhomogeneous group of attacks.
From \Cref{fig:papers-per-defence-type}, we can see that output perturbation and monitors are the most prominent defence techniques.
It should also be noted that some approaches such as monitoring and watermarking are reactive and thus aimed at \textit{detecting} an attack, while others are proactively trying to \textit{prevent} an attack.

\section{Background}
In this section, we briefly summarise important concepts, terminology and notation required for the rest of this paper.

\label{sec:background}
\subsection{Machine Learning}
So far, model stealing literature primarily targets \textit{supervised}- and \textit{reinforcement} learning.
In supervised learning, each sample $x_i$ from the input data set $X$ has a corresponding label $y$, and the goal is to learn a model that approximates the real mapping function $f(x)=y$ for a given problem $P$ to eventually predict labels $\hat{y}$ for unlabelled data. 
If the labels are discrete values, this is a classification problem; if they are continuous values, the problem is called regression.
In reinforcement learning, \textit{agents} are learning to make the best decision in a given situation so that the reward of a performed action is maximal. They act in a particular environment in order to achieve a predetermined goal. 

Successfully learning a Machine Learning model requires different resources and domain knowledge.
First, the quality of the model depends on the quality of the training dataset. This includes sample gathering and data labelling, often requiring human experts' knowledge which can be very resource-consuming.
Before learning the model, training hyperparameters, such as the learning rate for a Neural Network or the architecture of the model (e.g. the number of layers in a Neural Network), have to be set.
Selecting fitting values requires expert knowledge and experience.
Finally, model training itself can be very compute- and time consuming and may require many refinement cycles of hyperparameter setting and training in order to arrive at the most representative model. 
The need for large datasets, expert knowledge and compute resources are the main reasons for the emergence of the MLaaS paradigm.

In the following, we introduce Machine Learning concepts relevant for this paper.

In an \textbf{Active Learning} learning~\cite{settles_active_2009} process, an \textit{oracle} receives data samples and returns the corresponding labels.
Since data is labelled dynamically, one can choose the samples most useful to building the model, thus reducing the required amount of labelled data.
Thus, active learning is a strategy often employed to reduce the number of queries required during model stealing.
Generally, the "oracle" is considered to be a \textit{human (domain) expert} who is asked to provide an ad hoc ground truth; however, it can be any other information source.
In model stealing, the \textit{target model}, which can label samples, is thus considered to be the oracle.
This connection between active learning and model stealing has been explored by Chandrasekaran et al.~\cite{chandrasekaran_exploring_2020}; other works \cite{pal_framework_2019,pal_activethief_2020,pengcheng_query-efficient_2018,shi_active_2018} use active learning to improve attack efficiency.

\textbf{Knowledge Distillation}~\cite{bucilua_model_2006} is a model compression~\cite{cheng_model_2018} method that allows to train a smaller version (student network) of an already trained larger (teacher) network, without decreasing accuracy.
The main idea is that the student network is learning to duplicate the outputs of the teacher network on \textit{each}, and not only the final layer.
In model stealing, these ideas form the basis for some attacks (e.g. Kariyappa et al.~\cite{kariyappa_maze_2021}) as well as defences (e.g. Xu et al.~\cite{xu_deepobfuscation_2018}).

\textbf{Machine Learning as a Service} (MLaaS) refers to cloud-based computing platforms that offer Machine Learning tools. These services allow users to remotely train their models, evaluate them, or use pre-trained models via a \textit{pay-per-query} principle. 
Providers such as Amazon\footnote{\url{https://aws.amazon.com/machine-learning}}, Microsoft (Azure)\footnote{\url{https://azure.microsoft.com/en-in/services/machine-learning}}, or Google\footnote{\url{https://cloud.google.com/products/ai}} offer these services. 
Models supplied by MLaaS are usually only available for input-output interaction without revealing the model architecture and -parameters. 
If a model is trained on a cloud-based server by a user, its parameters and training hyperparameters may be revealed afterwards;
however, some MLaaS keep also the user's models secret, 
making it impossible to transfer the models to the user's device.
Amazon and Microsoft Azure e.g. provide two modes for model training: (1) A user does not specify training hyperparameters, but the server spends time searching optimal values for them. 
The MLaaS does not disclose these after the training. 
(2) Specified hyperparameters are required; therefore, it takes less time for running and costs less.
Wang et al.~\cite{wang_stealing_2018} have shown how to exploit the first mode for stealing the training hyperparameters.

We now briefly describe the methods most frequently targeted in model stealing attacks.
Naive Bayes (NB) applies Bayes’ theorem with the naive assumption of conditional independence between every pair of features. It uses the maximum a posteriori estimation to obtain the likelihood of a class for a given input.
A Decision Tree (DT) is a tree-structured model in which internal (decision) nodes represent conditions on the values of input features, branches represent the decision rules, and leaf nodes represent the outcome. 
If used for regression, the trees are called Regression Trees (RT).
Logistic Regression (LogReg) computes the odds of a class as a linear combination of the features and uses the logistic function to model a binary target variable; it can be extended to multi-class settings (MLogReg).
A Support Vector Machine (SMV) constructs a hyperplane that maximises the distance to the nearest training data point. For problems that are not linear separable, a kernel function maps the input samples into a higher-dimensional space, hoping that separation is possible there. 
Kernels include the linear (SVM-lin), quadratic (SVM-quad), or the radial basis function (SVN-RBF).
Some works subsume all linearly separating, binary-class models as \textit{linear binary model} (LBM).

\subsection{(Deep) Neural Networks and Deep Learning}
Many works in model stealing specifically address \textit{(Artificial) Neural Networks} ((A)NN) which consist of neurons that are organised in layers. 
The first is called the \textit{input layer}, the last the \textit{output layer}, and all in between are \textit{hidden layers}. The parameters of NNs usually are called \textit{weights}. 
Architectures frequently considered in model stealing research 
include:
\begin{itemize}
\item Deep Neural Networks (DNNs), i.e. neural networks with at least two hidden layers; often, these are fully-connected feed-forward networks: neurons from one layer can be connected only to neurons on the next layer.
 
\item \textit{Convolutional Neural Networks} (CNNs) are a special case of DNNs often applied to image data. They do not require feature extraction as a data preprocessing step, but can extract local spatial features in an end-to-end learning fashion \cite{goodfellow_deep_2016}; computationally, this feature extraction is relatively cheap. 
CNNs usually contain three types of layers: (1) \textit{Convolutional} layers apply filters to the layer's input and perform spatial feature extraction. (2) \textit{Pooling layers} are used for dimensionality reduction. (3) \textit{Fully connected} layers are performing the classification.

\item \textit{Recurrent Neural Networks} (RNNs) allow cyclic connections and support sequential data (of variable length), e.g. handwriting and speech recognition tasks. RNNs have an internal memory considering previous states.

\item \textit{Generative Adversarial Networks} (GANs) \cite{goodfellow_generative_2014} can be used to generate data; they consist of two networks competing with each other: the \textit{generator} learns to generate samples indistinguishable from the training samples, whereas the \textit{discriminator} learns to distinguish between original and generated data samples.

\item \textit{Graph Neural Networks} (GNN) process graph structures \cite{scarselli_graph_2009}, e.g. for social network analysis. GNNs can perform node classification, link prediction, or complete graph classification.
\end{itemize}

\subsection{Adversarial Machine Learning}
\label{sec:adversarial-ml}
Barreno et al.~\cite{barreno_can_2006} are amongst the first to explore security issues of Machine Learning and distinguish e.g. between attacks in the the model's training stage versus attacks on a trained model.
Biggio and Roli categorised adversarial attacks based on the attacker's goal and capabilities \cite{biggio_wild_2018} (see  \Cref{tab:adversarial}). 
The goal of an attack can be the model's confidentiality, integrity, or availability (the so-called "CIA triangle"). Confidentiality attacks are aimed at training data (e.g. model inversion) or the model as intellectual property (architecture and hyper(parameters)). 
Integrity attacks raise the number of false negatives. 
The goal of availability attacks is to make the model irrelevant by increasing prediction errors.

\begin{table*}[t]
\centering
\caption{Attacks against Machine Learning, adapted from \cite{biggio_wild_2018}}
\label{tab:adversarial}
\newcolumntype{L}[1]{>{\raggedright\let\newline\\\arraybackslash\hspace{0pt}}m{#1}}

\centering
\begin{tabular}{|lp{2cm}|L{3.6cm}L{2.4cm}L{4.7cm}|}
\hline
     & \small{\textbf{Attacker's }}                                    & \multicolumn{3}{c|}{\small{\textbf{Attacker's goal}}}                                                                                                                                       \\ \cline{3-5}
\multicolumn{2}{|c|}{\small{\textbf{capability}}} & \small{\textbf{Integrity}}                                                                   & \small{\textbf{Availability}}                      & \small{\textbf{Privacy/Confidentiality}}               \\ \cline{1-5} 
     & \multicolumn{1}{l|}{\small{\textbf{Test data}}}      & \small{Evasion (e.g., adversarial examples)}                                                 & \small{-}                                          & \small{Model extraction/stealing, model inversion, membership inference, \dots} \\ \cline{1-5} 
     & \multicolumn{1}{l|}{\small{\textbf{Train data}}}     & \small{Poisoning for subsequent intrusions - e.g., backdoors} & \small{Poisoning to maximise error} &  \small{-}                                  
     \\ \hline
\end{tabular}

\end{table*}

\begin{itemize}
\item \textit{Poisoning} attacks \cite{nelson_exploiting_2008} poison the training data, for instance by flipping the labels or adding some malicious data into the training set. As a consequence, the trained model's accuracy is lower or it can be fooled by samples modified in the same manner as the training data. 

\item \textit{Evasion} attacks target the prediction phase. By e.g. applying small perturbations to original data \cite{szegedy_intriguing_2014}, an adversary can obtain an adversarial example that is most of the time indistinguishable for humans, but misclassified by the model. These attacks have become prominent for images, but were first executed on email (i.e. text data) \cite{lowd_good_2005}.

\item \textit{Membership inference} attacks \cite{shokri_membership_2017} determine if a given sample belongs to the training data or not. To do this, an adversary tries to distinguish the differences in the predictions of inputs in the training set and outside of it.

\item \textit{Model stealing} (model extraction) reveals a model's hyperparameters resp. learned parameters or steals model behaviour and, thereby, the intellectual property a model constitutes. Model stealing is the focus of this work.
\end{itemize}

\section{Taxonomy of Model Stealing Attacks}
\label{sec:model_stealing}
In this section, we first provide a unified terminology (\Cref{sec:terminology}), followed by a comprehensive taxonomy of model stealing attacks (\Cref{sec:taxonomy}).

\subsection{Terminology and Notation}
\label{sec:terminology}
\begin{table*}[t]
\centering
\caption{Disambiguation of model stealing terminology. The first column gives the term primarily used in the literature and, thus, also in this paper. The second column lists other, equivalent terms used across the literature.}
\label{tab:terminology}
\newcolumntype{L}[1]{>{\raggedright\let\newline\\\arraybackslash\hspace{0pt}}m{#1}}
\begin{tabular}{|L{2.8cm}|L{11.5cm}|}
\hline
\small{\textbf{Terminology used in this paper}} & \small{\textbf{Other designations with the same meaning}} \\ \hline

\small{Model stealing (extraction) attack \cite{tramer_stealing_2016}} & \small{Reverse-engineering attack \cite{lowd_adversarial_2005}, 
copy attack \cite{correia-silva_copycat_2018}, exploratory attack \cite{yi_shi_how_2017}, inference attack \cite{shi_generative_2018}, duplication attack \cite{joshi_gdalr_2019}, mimicking attack \cite{mosafi_stealing_2019}, model approximation attack \cite{ali_best-effort_2020}} \\ \hline

\small{Target model \cite{tramer_stealing_2016}}            & \small{(Target) oracle \cite{tramer_stealing_2016, pengcheng_query-efficient_2018}, classifier (model) under attack \cite{yi_shi_how_2017}, secret model \cite{pal_activethief_2020}, victim model \cite{orekondy_knockoff_2019}, 
original model \cite{reith_efficiently_2019}, proprietary model \cite{chandrasekaran_exploring_2020}, mentor model \cite{mosafi_stealing_2019}, source model \cite{lukas_deep_2021}}  \\ \hline

\small{Substitute model \cite{papernot_practical_2017}}     & \small{Adversarial classifier (model) \cite{yi_shi_how_2017}, copycat network \cite{correia-silva_copycat_2018}, knockoff model \cite{orekondy_knockoff_2019}, surrogate model \cite{atli_extraction_2020}, extracted model \cite{reith_efficiently_2019}, inferred classifier (model) \cite{shi_generative_2018}, model approximation \cite{chandrasekaran_exploring_2020}, student model \cite{mosafi_stealing_2019}, stolen model \cite{yuan_es_2022}, replicated model \cite{chen_stealing_2021}, clone model \cite{miura_megex_2021}} \\ \hline

\small{Attacker's data} & \small{Fake dataset \cite{correia-silva_copycat_2018}, thief dataset \cite{pal_activethief_2020}, attacker set \cite{juuti_prada_2019}, transfer set \cite{orekondy_knockoff_2019}, proxy data \cite{barbalau_black-box_2020}, surrogate dataset \cite{truong_data-free_2021}}    \\ \hline

\small{Fidelity \cite{jagielski_high_2020}}                 & \small{Extraction accuracy \cite{tramer_stealing_2016}, label prediction match \cite{papernot_practical_2017}, similarity \cite{pengcheng_query-efficient_2018}, agreement \cite{pal_activethief_2020}, approximation accuracy \cite{reith_efficiently_2019}} \\ \hline

\end{tabular}

\end{table*}
We present a unified terminology in \Cref{tab:terminology}. We identify the most widely used terms in the literature in the first column and adhere to them in our paper. The second column indicates alternative terms along with a list of works that utilise them.
A model that an adversary aims to steal is called the \textit{target model} and is denoted as $f$. The adversary can use this model as an oracle to collect the \textit{attacker's data} that consists of pairs $(x, y)$. The input $x$ is a data sample that the attacker sends to the oracle. The output $y$ is the prediction of the target model, i.e. $f(x)=y$. One such interaction with the target model is called a \textit{query}. 
If outputs are the only information one can obtain from $f$, we assume that an adversary has \textit{black-box} access to the target model, or that $f$ is a \textit{black box}. If the architecture and parameters of the target model are known, we assume \textit{white-box} access to the model, or that $f$ is a \textit{white box}. Any in-between state is called \textit{grey box}.
If the attacker obtains a (possibly approximate) copy of the target model, we denote that model with $\hat{f}$.

\begin{figure*}[t]
\includesvg[width=0.8\textwidth]{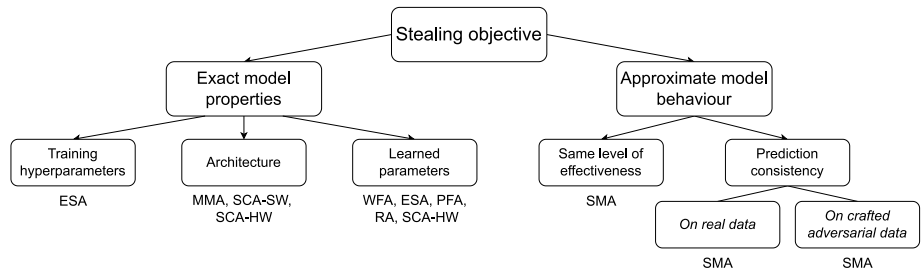}
\centering
\caption{Taxonomy of model extraction attacks.} 
\label{fig:taxonomy}
\end{figure*}

\subsection{Objectives of Model Stealing Attacks} \label{sec:taxonomy}
By their objective, as depicted in \Cref{fig:taxonomy}, attacks can be divided into two categories: (1) stealing \textit{exact model properties} (\Cref{sec:stealing_objectives}), and (2) stealing \textit{approximate model behaviour} (\Cref{sec:performance_objectives}).

\subsubsection{Stealing Exact Model Properties}
\label{sec:stealing_objectives}

Depending on the considered task, the stealing of exact properties can further be distinguished by the stolen \textit{assets}: the learned parameters (e.g. the learned weights of a neural network), training hyperparameters (e.g. a regularisation parameter utilised during training), or architecture (e.g. the arrangement of nodes and layers in a neural network).

\textbf{Training Hyperparameters.}
In this category of attacks, an adversary tries to reveal a hyperparameter responsible for the training process. Wang and Gong \cite{wang_stealing_2018} proposed an equation-solving approach for stealing the regularisation hyperparameter from ridge and logistic regression, SVM and NN (see \Cref{sec:ESA}). Oh et al. trained a \textit{meta-model} that can predict some of the training hyperparameters, such as batch size or optimisation algorithm \cite{oh_towards_2018} (c.f. \Cref{sec:meta-model}).

\textbf{Architecture.}
An architecture stealing attack is usually applied to neural networks, as most other models vary only on training hyperparameters and have a fixed architecture. In this case, "architecture" means the set of hyperparameters that defines the target model structure. In particular, the number of layers, layer type and its characteristics like the size of a kernel are parts of a CNN architecture.
Two main approaches to architecture stealing have been proposed in the literature. The first one is the aforementioned meta-model attack \cite{oh_towards_2018} that predicts the architecture of the target model by utilizing queries.
Other works \cite{hua_reverse_2018,hu_deepsniffer_2020,yan_cache_2020,hong_security_2020,xiang_open_2020,zhu_hermes_2021} exploit side-channel access to the model. We provide a taxonomy of side-channel attacks, describe the difference from query-based attacks, and introduce key techniques in \Cref{sec:side-channel}. 

\textbf{Learned Parameters.}
A parameter stealing attack aims to extract parameters of the target model whose structure (architecture) is known. Lowd and Meek were the first to propose a model extraction attack for stealing the weights of a binary classifier \cite{lowd_adversarial_2005}. Later, Tramèr et al.~\cite{tramer_stealing_2016} extended the idea and introduced equation-solving attacks (see \Cref{sec:ESA}) which allow to extract the exact parameters of 
(multi-class) logistic regression and Multi-Layer Perceptron. Reith et al. \cite{reith_efficiently_2019} presented an equation-solving attack also for support vector regression with linear- or quadratic kernels.
Generally, learned parameter extraction is highly related to other attack types. For instance, it can be applied after an architecture stealing attack to steal a target model with an unknown model type. By performing a successful extraction of model parameters, an adversary automatically obtains \textit{identical} behaviour. 

\subsubsection{Stealing Model Behaviour}
\label{sec:performance_objectives}

Jagielski et al. \cite{jagielski_high_2020} classified a subset of those attacks that aim to steal the model behaviour based on their accuracy- and fidelity performance. We generalise from concrete metrics to the goals of obtaining the \textit{same level of effectiveness} as the target model, or trying to be \textit{consistent with the predictions} of the target model; we further distinguish two cases for the latter, depending on what they are tested on.
In \Cref{sec:goals}, we detail concrete, frequently employed metrics to measure these goals, and in \Cref{tab:attacksPerformance-non-SMA,tab:attacksPerformance-SMA}, we analyse model stealing attacks based on their performance objectives. 

\textbf{Same Level of Effectiveness.}
This category covers attacks that aim at approximate stealing and that focus on getting a copy of the target model that reaches the same level of effectiveness.
Given the target model $f$, an attacker aims to create a model $\hat{f}$ that performs similarly to $f$ on the original data. As a result, the attacker can use $\hat{f}$ for solving the same task as $f$ without restrictions (e.g. daily caps or fees). 
To get a stolen model with similar effectiveness performance, an adversary can use the same model architecture as in the target model \cite{tramer_stealing_2016,chandrasekaran_exploring_2020,correia-silva_copycat_2018}, the same model type but with a different structure \cite{orekondy_knockoff_2019,shi_active_2018}, or use a completely different class of models \cite{yi_shi_how_2017,teitelman_stealing_2020}.  

\textbf{Prediction Consistency.}
The second category that aims at approximate behaviour stealing covers attacks that produce a model $\hat{f}$ that predicts outputs consistently with $f$. Consistent predictions means that for any sample $x$, the stolen model prediction should coincide with the target model prediction $f(x) = \hat{f}(x)$. Hence, if $f$ misclassifies a sample from the original data, we want $\hat{f}$ to also misclassify it. Depending on the domain of $x$, we distinguish two sub-categories: consistency on real data and consistency on crafted adversarial data. For the real-data case, we will get a model that has the same effectiveness as the target model and makes the same mistakes, e.g. on the original data. In that sense, $f$ and $\hat{f}$ are more similar than in the previous category. 
Prediction consistency on crafted adversarial data can be the goal of a model stealing attack which opens the black-box target model for further white-box attacks.

\section{Model Stealing: Threat Model}\label{sec:modelStealing}

A common way to understand the mechanics of a security attack is to model potential threats. Hence, we specify the attacker's motivation (incentives) to perform a model stealing attack and describe potential consequences for the model owner. Then we analyse how the attacker may execute an attack. This step is also helpful for modelling potential defences as it reveals the attack's weaknesses. Finally, we formulate concrete goals for model stealing attacks and define metrics to measure the level of success.

\subsection{Attacker's Incentives}
\label{sec:incentives}
We distinguish the following two reasons for a model stealing attack. 

\textbf{Exploit a (partial) copy of the target model.} If the target model is only available via API as a black box, there might be some restrictions that prevent API users from unlimited model querying -- for instance, daily caps or fees. If the attacker wants to overcome these, obtaining a copy of the model would be the solution. Another motivation is e.g. stealing a novel architecture, which could help an adversary to get a better model for another task (i.e. not necessarily the one that the target model solves). In this case, the attacker does not steal the model itself, but one of its components. 

\textbf{"Open" the target model for further white-box attacks.} An attacker may want to perform an attack that requires white-box access to the target model. In this case, a model stealing attack can be used as an intermediate step. By obtaining a copy of the target, the attacker "opens" the black box to perform some white-box attack on it. 
Several works exploit model stealing to enable evasion- or poisoning attacks, for example \cite{shi_evasion_2017,sethi_security_2017,sethi_data_2018}. As mentioned in \Cref{sec:methodology}, we did not include these papers in our classification since -- in contrast to those mentioned in \Cref{tab:attacks} -- they are focused on the subsequent white-box attacks; here, model stealing is not actually studied, but rather used as a preparatory step.

\subsection{Attacker's Capabilities}
\label{sec:capabilities}
We consider three main aspects regarding the attacker's capabilities: knowledge about the target model and the data it was trained on (the \textit{original} data), actions that the attacker can perform, and the resources available to them.  

\textbf{Attacker's Knowledge.} As mentioned in \Cref{sec:terminology}, an attacker might have one of the three types of access to the target model: white-box, grey-box, or black-box. Having white-box access means that there is no reson for stealing the model as it is already known. 
However, such a model can still be exploited to reveal its training hyperparameters \cite{wang_stealing_2018}. Grey-box access refers to the situation when the architecture of the target model is known, which is required for some types of attacks (see \Cref{sec:query-based},  \Cref{tab:attacksPerformance-non-SMA,tab:attacksPerformance-SMA} for more details). 
However, the default assumption for model stealing attacks is that there is only black-box access, i.e. the only information revealed are model outputs. 
Some of the model stealing attacks are data-agnostic, not requiring any data at all (see \Cref{sec:side-channel}) or at least no meaningful data (see, for instance, \Cref{sec:ESA}). However, there are many attacks (\Cref{sec:attack:query:substitute}) for which the quality of the data is important. We focus on the following categories: original data, problem-domain data, non-problem domain data, and artificial data. We describe these categories in detail in \Cref{sec:sma-data}.

\textbf{Attacker's  Actions.} \textit{Queries} are the basic interactions between the attacker and the target model (cf. \Cref{sec:terminology}). We call attacks that use only this type of action as information source \textit{query-based} attacks. These attacks are therefore suitable in an MLaaS setting. An overview of query-based attacks is presented in \Cref{sec:query-based}.
If the attacker has hardware- or software access to the computing resource on which the model is deployed, this opens an additional possibility for a model stealing attack. In these settings, the attacker can exploit \textit{side-channel leakages}, thus performing so-called \textit{side-channel attacks} (SCAs). We present an overview of side-channel attacks in \Cref{sec:side-channel}. SCAs can optionally also use queries as an additional source of information. 

\textbf{Attacker's Resources.} As discussed in \Cref{sec:incentives}, there might be restrictions that affect the number of queries an attacker can perform. Hence, (query-based) model stealing attacks are usually considered with regards to their \textit{query budget} -- i.e. the number of queries that an attack requires. It is an important task to find a trade-off between the attack performance and its query budget.

Based on their capabilities, we can call attackers \textit{weak} or \textit{strong}. For instance, an adversary who knows about the architecture of the target model and the original training data is stronger than one without that knowledge. It is not always possible to say which capabilities make a stronger attacker (i.e less knowledge and more resources v. more knowledge and fewer resources). However, we can differentiate within the categories knowledge, actions, and resources. 

\subsection{Attacker's Goals}
\label{sec:goals}

Every model stealing attack aims to copy the target model or some of its aspects. As discussed in \Cref{sec:model_stealing},
on the top level, we separate the attacks into two categories: (1) stealing exact model properties, and (2) stealing approximate model behaviour (\Cref{fig:taxonomy}). 
In this section, we focus on defining metrics which estimate if a certain goal was reached.

\textbf{Effectiveness.}
Depending on the stealing objective, the effectiveness of the attack is measured differently.
\begin{itemize}
    \item 
    Effective \textit{exact model properties extraction} means that the extracted values are very close or equal to the corresponding target values. Thereby, the most common way to measure the effectiveness is to calculate the absolute difference between target- and stolen values. 
    
    \item 
    The effectiveness of model stealing attacks that \textit{aim to steal behaviour} is usually measured with one or several metrics; below, we define accuracy, fidelity and transferability and describe their relevance for model stealing attacks. Additionally, different error rates can be calculated, but they are inverse to the metrics mentioned above; hence, we do not focus on them.     

        \begin{itemize}
    
        \item 
        \textit{Accuracy} shows how close model predictions are compared to the ground-truth values. It is calculated on both target- and stolen models, and results are expected to be similar. However, even equal performance does not mean that the stolen model simulates the original model perfectly -- models can still yield different predictions for single data points, and averaged identical accuracies can just be a coincidence. 
        This metric is used to evaluate approximate stealing attacks which aim to reach the same level of effectiveness as the target model.

        \item 
        \textit{Fidelity} is calculated as the accuracy of the substitute model when the target model predictions are considered the ground-truth. This metric shows how well the stolen model simulates the original. Furthermore, fidelity does not require ground-truth labelling since it uses only the labels of the target model, which can be observed through querying the model. Consequently, it can be calculated on any data from any distribution without losing its relevance. 
        Fidelity can be used to evaluate the success of an attack which aims to create a model that consistently makes the same predictions as the target model.

        \item 
        \textit{Transferability} shows how many adversarial examples generated for the stolen model $\hat{f}$ are also adversarial for the target model $f$. In other words, let $x$ be a real data sample, $f(x) = \hat{f}(x) = y$, and $x^*$ be an adversarial example for $\hat{f}$, so $\hat{f}(x) \neq \hat{f}(x^*)$. Having $f(x) \neq f(x^*)$ then means that there is transferability between the stolen model and the target model. To measure transferability numerically, one can create a test set of adversarial examples crafted for $\hat{f}$ and measure how many of them are  misclassified by $f$.
        This metric is used when an adversary wants to reach high prediction consistency on crafted adversarial data for, e.g., targeting a black-box model with an evasion attack. 
        Papernot et al. \cite{papernot_transferability_2016} showed that adversarial examples, crafted via exploiting a stolen black-box model, lead to a high misclassification rate on the target model.
        \end{itemize}
\end{itemize}

\textbf{Efficiency.}
To measure the efficiency of an attack, two metrics are usually used: the number of queries, i.e. the \textit{query budget}, and the \textit{time} needed to carry out the attack. We focus on \textit{the number of queries per parameter} which is a metric often employed for analysing equation-solving attacks (\Cref{sec:ESA}), but is relevant for most query-based attack types.

\begin{itemize}
    \item 
    The \textit{number of queries} (query budget) corresponds to the price an adversary pays for performing an attack, and is usually calculated only for query-based attacks. 
    A drawback of this metric is that it can only be compared for models of the same size, as the amount of required queries generally increases with the learned parameters

    \item 
    To account for that, the \textit{number of queries per parameter} is calculated as the query budget divided by the number of learned parameters of the target model, and thus allows to compare attacks across different target model.
    
    \item 
    \textit{Timing} is measured less frequently, and it usually means the time of preparation for an attack. For instance, if an API provides 1,000 queries per day for free and an adversary has no budget but wants to apply an attack that requires 3,000 queries, three days will be spent just on data collection. 
    Since this metric can depend on API- or computational resources of an attacker, we do not consider it in this paper (cf. \Cref{tab:attacksPerformance-non-SMA,tab:attacksPerformance-SMA}).
\end{itemize}

\section{Query-based Model Stealing Attacks}
\label{sec:query-based}

In this section, we discuss query-based model stealing approaches. We group them by the stealing method and analyse each of the methods separately. In particular, we describe the adversary's capabilities for each attack and analyse their efficiency and effectiveness. \Cref{tab:attacks} presents the taxonomy of query-based attacks, using the categorisation described in \Cref{sec:stealing_objectives}. If a paper only proposed an improvement of a known attack and does not perform the attack itself, we did not include it (e.g. \cite{joshi_gdalr_2019}). 
\begin{table*}[ht]
\centering
\caption{Taxonomy of query-based attacks. Note that the attack goal of stealing learned parameters in most cases also implicitly provides behaviour stealing.}
\label{tab:attacks}
\resizebox{\textwidth}{!}{\newcolumntype{L}[1]{>{\raggedright\let\newline\\\arraybackslash\hspace{0pt}}m{#1}}
\begin{tabular}{|L{2.5cm}|L{2.5cm}|L{2.5cm}|L{4cm}|L{1.7cm}|}
\hline
\small{\textbf{Attack goal}}                      & \small{\textbf{Stealing method}} & \small{\textbf{Data domain}}                                       & \small{\textbf{Target model}}                                                          & \small{\textbf{Papers}}                                                                                                                                                                                                                                                                                                                                                                                                                                                                                                                                                                                                                                                                                                              \\ \hline
\multirow{2}{2.5cm}{\small{Training hyperparameters}} & \small{Meta-model}               & \small{Image}                                                      & \small{DNN, CNN}                                                                       & \small{\cite{oh_towards_2018}}                                                                                                                                                                                                                                                                                                                                                                                                                                                                                                                                                                                                                                                                                                       \\ \cline{2-5} 
                                                  & \small{Equation-solving}         & \multirow{4}{*}{\small{Tabular}}                                   & \small{RR, LR, SVM, NN}                                                                & \small{\cite{wang_stealing_2018}}                                                                                                                                                                                                                                                                                                                                                                                                                                                                                                                                                                                                                                                                                                    \\ \cline{1-2} \cline{4-5} 
\multirow{4}{*}{\small{Learned parameters}}       & \small{Witness-finding}          &                                                                    & \small{LBM, SVM-poly}                                                                  & \small{\cite{lowd_adversarial_2005, tramer_stealing_2016, reith_efficiently_2019}}                                                                                                                                                                                                                                                                                                                                                                                                                                                                                                                                                                                                                                                   \\ \cline{2-2} \cline{4-5} 
                                                  & \small{Equation-solving}         &                                                                    & \small{LR, MLR, MLP, SVR-lin/quad}                                                     & \small{\cite{tramer_stealing_2016, reith_efficiently_2019, yan_monitoring-based_2021}}                                                                                                                                                                                                                                                                                                                                                                                                                                                                                                                                                                                                                                               \\ \cline{2-2} \cline{4-5} 
                                                  & \small{Path-finding}             &                                                                    & \small{DT, RT}                                                                         & \small{\cite{tramer_stealing_2016}}                                                                                                                                                                                                                                                                                                                                                                                                                                                                                                                                                                                                                                                                                                  \\ \cline{2-5} 
                                                  & \small{Recovery}                 & \multirow{3}{*}{\small{Image}}                                     & \small{ReLU-DNN}                                                                       & \small{\cite{jagielski_high_2020, milli_model_2019, rolnick_reverse-engineering_2020, carlini_cryptanalytic_2020}}                                                                                                                                                                                                                                                                                                                                                                                                                                                                                                                                                                                                                   \\ \cline{1-2} \cline{4-5} 
\multirow{2}{*}{\small{Architecture}}             & \small{Meta-model}               &                                                                    & \small{DNN, CNN}                                                                       & \small{\cite{oh_towards_2018}}                                                                                                                                                                                                                                                                                                                                                                                                                                                                                                                                                                                                                                                                                                       \\ \cline{2-2} \cline{4-5} 
                                                  & \small{Recovery}                 &                                                                    & \small{ReLU-DNN}                                                                       & \small{\cite{rolnick_reverse-engineering_2020}}                                                                                                                                                                                                                                                                                                                                                                                                                                                                                                                                                                                                                                                                                      \\ \hline
\small{Level of effectiveness}                    & \small{Substitute model}         & \small{Image, Tabular, Text, Sequential, Graph, RL environment}    & \small{LBM, MLogReg, DT, RF, SVM, NN, CNN, BERT, DRL, chip, GNN, GAN, Encoder}         & \small{\cite{orekondy_knockoff_2019, correia-silva_copycat_2018, atli_extraction_2020, chandrasekaran_exploring_2020, roberts_model_2019, mosafi_stealing_2019, yuan_es_2022, kariyappa_maze_2021, aivodji_model_2020, krishna_thieves_2020, takemura_model_2020, teitelman_stealing_2020, wu_model_2022, he_stealing_2021, milli_model_2019, barbalau_black-box_2020, yu_cloudleak_2020, chen_stealing_2021, szyller_good_2021, he_model_2021, hu_stealing_2021, truong_data-free_2021, miura_megex_2021, zhang_thief_2021, aarts_leveraging_2021, wang_enhance_2022, liu_stolenencoder_2022, sha_cant_2022, wang_black-box_2022, yan_towards_2022, sanyal_towards_2022, xie_game_2022, dziedzic_difficulty_2022, shen_model_2022}} \\ \hline
\small{Prediction consistency}                    & \small{Substitute model}         & \small{Image, Tabular, Text, Graph, RL enviroment, Recommendation} & \small{(M)LogReg, kNN, DT, LGBM, SVM, SVR, NB, NN, CNN, RNN, BERT, DRL, GNN, GAN, SRS} & \small{\cite{tramer_stealing_2016, papernot_practical_2017, yi_shi_how_2017, reith_efficiently_2019, shi_active_2018, shi_generative_2018, pengcheng_query-efficient_2018, pal_activethief_2020, juuti_prada_2019, yuan_es_2022, aivodji_model_2020, krishna_thieves_2020, behzadan_adversarial_2019, defazio_adversarial_2020, wu_model_2022, pal_framework_2019, ali_best-effort_2020, gong_inversenet_2021, yue_black-box_2021, hu_stealing_2021, wang_black-box_2022, xie_game_2022, shen_model_2022, wang_dualcf_2022}}                                                                                                                                                                                                         \\ \hline
\end{tabular}}
\end{table*}
If authors claim their attack to be a behaviour stealing attack, but actually provide a method for stealing parameters, we define their goal as parameter stealing in \Cref{tab:attacks}. Such classification does not contradict the one defined by the authors since a high-performing parameter stealing attack usually leads to a model with equivalent behaviour. 
We define the goal of an attack as the "level of effectiveness" if the performance of the attack is evaluated using accuracy. If the performance is evaluated using fidelity or transferability, the goal of the attack is defined as "prediction consistency". In some of the papers both accuracy and fidelity (or accuracy and transferability) were measured. Those papers belong to both categories simultaneously. 
Success measures for these methods are discussed below in \Cref{tab:attacksPerformance-non-SMA,tab:attacksPerformance-SMA}; this section focuses on the description of the approaches.

\subsection{Witness-finding Attack}
Lowd and Meek presented the earliest model stealing attack which aims to steal parameters of linear binary models (LBMs) \cite{lowd_adversarial_2005}. Considering one positive and one negative sample, they changed the feature values of the positive sample one by one until they found \textit{sign witnesses} - i.e. a couple of samples which are identical except one feature value $f$ and belong to different classes. They set the corresponding weight $w_f$ to $1$ or $-1$ depending on sign witness values and then used a line search to reveal the relative weight of other features. Since the main step of the attack is to find sign witnesses, we call it the \textit{witness-finding attack} (WFA).
An adversary needs the target model architecture and two data samples (one positive and one negative) to perform this attack. Since the attack allows the exact extraction of weights, it produces a model with the same performance as the target model. The main drawback is the inefficiency of the attack: it takes at least 11 queries per parameter (weight) to steal a model (cf. \Cref{tab:attacksPerformance-non-SMA}), which can be problematic for large models.  
More than a decade later, Tramèr et al. and Reith et al. adapted the witness-finding attack to Support Vector Machines (SVMs) and Support Vector Regression Machines (SVRs) \cite{tramer_stealing_2016, reith_efficiently_2019}.

\subsection{Equation-solving Attacks}
\label{sec:ESA}
We define an attack as \textit{equation-solving attack} (ESA) if it is based on setting up a system of equations and solving it. The solution corresponds to the values an adversary wants to extract. Thereby, an ESA appears when the extraction goal is the exact values of the target model -- more specifically, either the learned parameters or the training hyperparameters.

ESAs were first utilised by Tramèr et al. for stealing learned parameters of (Multi-class) Logistic Regression, and Multi-Layer Perceptron \cite{tramer_stealing_2016}. They sent data samples $x_1, \ldots, x_n$ to the target model $f_w$ with learned parameters $w$ and used the outputs $y_1, \ldots, y_n$ to construct the system of equations $f_w(x_i) = y_i$, $i=1, \ldots, n$.
The solution of the system reveals the values of the parameters $w$. 
Similarly to the witness-finding attack, it reaches perfect extraction scores (cf. \Cref{tab:attacksPerformance-non-SMA}). However, an ESA is more efficient, requiring $1$ to $4$ queries per parameter depending on the target model type. The attack also requires the architecture of the target model to be known and data samples to query the model. However, since queries are only used to construct a system of equations, the attacker can use samples which are not necessarily real or meaningful. 
Reith et al. applied this attack for stealing parameters of SVRs with linear or quadratic kernels \cite{reith_efficiently_2019}. 
Yan et al. used an ESA to steal an MLP under a differential privacy defence \cite{yan_monitoring-based_2021} which adds noise to the outputs close to the decision boundary \cite{zheng_bdpl_2019} (cf. \Cref{sec:defences:prevention:perturbation}). They duplicated queries and, by observing different outputs for the same inputs, created a system of equations, the solution to which approximates the outputs of the target model.

Wang and Gong used an ESA to steal a regularisation hyperparameter, used in the objective function to balance between a loss function and a regularisation term \cite{wang_stealing_2018}. The learned parameters of the target model should minimise the value of the objective function. Hence, the gradient of the objective function, calculated on the model's parameters, should be (close to) $0$. Based on this, an adversary first computes gradients of the objective function and sets them to $0$. An obtained over-determined system can be solved by using the linear least square method.
To perform this attack, an adversary needs white-box access to the target model. Thereby, for extracting training hyperparameters with having only black-box access, one must first perform an architecture and learned parameters extraction attack.

\subsection{Path-finding Attacks}
The \textit{path-finding} attack (PFA) was presented by Tramèr et al. for stealing Decision Trees (DTs) and Regression Trees (RTs) \cite{tramer_stealing_2016}. This attack requires prediction labels and an identifier of the leaf that outputs the label.
An adversary sends an input $x$ to the target tree $t$ and collects an output $t(x)$ and a leaf identifier $id$. Then, by varying values of features, the adversary uncovers the conditions that an input sample has to satisfy in order to reach the leaf $id$. 
The leaf identifier is required to be able to distinguish between different leaf nodes that lead to the same label being returned, and thus would be indistinguishable by that information alone.
Besides the predictions, this attack also recreates the conditions a sample has to satisfy in order to be classified by a specific leaf.
If all leafs have unique ids, the stolen tree has identical behaviour to the original one. For regression trees, the authors achieved perfect fidelity scores as all leaves had unique identifiers. For classification tasks, this was not the case and the performance was worse (cf. \cref{tab:attacksPerformance-non-SMA}). Compared with the aforementioned attacks, the PFA is the most inefficient, requiring 66-317 queries per parameter depending on the target tree. The authors tried to optimise the attack by allowing samples with only some of the features, so-called "non-complete queries". As a result, the number of queries per parameter decreased to 44-91.

\subsection{Recovering Attacks}
\label{sec:recovering}
\textit{Recovering attacks} (RAs) are designed to reveal the weights or even the architecture of (D)NNs with (at least partially) linear activation functions. All works we identified focus on (D)NNs with ReLU activation functions (ReLU-(D)NNs). Milli et al. were the first who theoretically described a recovering attack for stealing parameters of ReLU-(D)NNs with two layers \cite{milli_model_2019}. They claimed that the ReLU network's weights could be viewed as separating hyperplanes. By finding input points that lie on these hyperplanes, one can recover the weights up to their signs. These points are also called critical- \cite{jagielski_high_2020,carlini_cryptanalytic_2020} and boundary \cite{rolnick_reverse-engineering_2020} points. Finally, one can recover the sign of the weights by querying samples and solving a system of equations.
Jagielski et al. implemented such an attack and showed that the stolen model has very high fidelity and perfect transferability \cite{jagielski_high_2020} (see \Cref{tab:attacksPerformance-non-SMA}). 
Rolnick et al. extended this approach for stealing a model of arbitrary depth and revealing its architecture \cite{rolnick_reverse-engineering_2020}.
Carlini et al. compared the model stealing problem with crypto-analysis of block ciphers \cite{carlini_cryptanalytic_2020}. They proposed an attack that reveals weights of ReLU networks and requires fewer queries than Rolnick et al. (see \Cref{tab:attacksPerformance-non-SMA}).
RAs are the least efficient for small models among the attacks presented in \Cref{tab:attacksPerformance-non-SMA}. It takes approximately $312$ queries per parameter to steal a model with $210$ parameters \cite{carlini_cryptanalytic_2020}. For models with tens of thousands of parameters, this score decreases to $12$, still leaving these attacks among the most inefficient ones. However, since a successful RA leads to the exact copy of the target model, it fully reproduces its behaviour.  

\subsection{Substitute Model Training} \label{sec:attack:query:substitute}
This approach has been widely used over the past years by numerous authors (see \Cref{tab:attacks} and \Cref{fig:papers-per-attack-type}). The idea is to train a substitute model (cf. \Cref{tab:terminology} for alternative terminology used throughout the literature, e.g. "surrogate model") using data labelled by the target model, i.e. using the target model as an oracle for the labels.
The substitute model can have the same architecture as the target model; however, this is not necessary and usually not the case. The main condition is rather a syntactical input-output correspondence between models, i.e. the substitute model has to accept the same format of inputs and return outputs in the same representation as the target model.

The first substitute model attack, besides other attacks presented, was introduced by Tramèr et al. \cite{tramer_stealing_2016}. They called the attack "retraining", but in fact they trained a substitute model for LBMs, (multi-class) Logistic Regression, MLP and SVM with RBF kernel, assuming that an adversary knows which target model is used. Reith et al. used the same approach for stealing SVMs and SVRs with linear and quadratic kernel \cite{reith_efficiently_2019}.
Another early form of this attack was presented by Papernot et al. \cite{papernot_practical_2017}. They proposed two substitute models: a more complex DNN and the simpler LogReg for stealing DNN, LogReg, SVM, Decision Trees, and k-NN. The main goal was to train a model with a decision boundary similar to the one of the original model in terms of transferability, i.e. that the an approximation of the original model allows to craft adversarial examples that will, with high probability, fool the target model. The substitute model can thus be used to carry out evasion attacks against the target model.

In the following, we discuss different aspect to substitute model training, namely (i) the substitute model architecture, (ii) the type and domain of training data, and (iii) the strategies to pick the samples and thus required number of queries.
Attacks against specific model types and domain-specific models are discussed in \Cref{sec:substitute:domainspecific}.

\subsubsection{Substitute Model Architecture}
In general, to perform a substitute model attack an adversary first has to pick a model's architecture. Usually, this decision is influenced by the type of model inputs. For instance, if a model is an image classifier, a CNN can be a good choice. Recent works have shown that for a higher stealing success rate, an adversary's model has to be at least as deep (complex) as the target model  \cite{yi_shi_how_2017, juuti_prada_2019, orekondy_knockoff_2019, krishna_thieves_2020, takemura_model_2020}; this applies also to language models and LSTMs \cite{krishna_thieves_2020, takemura_model_2020}.
Shi et al. experimented with stealing Naive Bayes and SVM using DNNs, and vice versa \cite{yi_shi_how_2017}. Their results also showed that using a more complex model (DNN) results in a better-performing substitute model.

\subsubsection{Substitute Model Training Data}
\label{sec:sma-data}
Another important aspect of substitute model training attacks is the dataset used for training. This dataset is often unlabelled, and thus, the target model is first queried with it to observe corresponding labels. The data and obtained labels then form the training data for the substitute model. 
We can distinguish various scenarios regarding the domain of the problem or data. 
While in regards to model stealing a clear definition of "domain" is lacking, this concept has been explored in detail in the context of Transfer Learning\cite{pan_survey_2010} where "domain" is characterised as consisting of two components \cite{pan_survey_2010}: the feature space and a marginal probability distribution. If two domains are different, then they may differ in feature space or the marginal probability distribution. We will use these components to characterise different settings in model stealing.
As described in \Cref{sec:capabilities}, we distinguish four categories of data: original, problem domain, non-problem domain, and artificial. 

\textbf{Original data} is the data that was actually used to train the target model. While some attacks assume the availability of this data, and it corresponds to an attacker with the strongest data knowledge, it may not be a realistic scenario.

\textbf{Problem Domain (PD) data} \cite{papernot_practical_2017} is data drawn from a distribution that closely resembles the original dataset -- for example, using images depicting human faces to steal a model trained for face recognition.
This would be data where the feature space is the same, and the marginal probability distribution might be quite similar, but not identical. 
In most cases, this data is obtained from public data repositories. 
Depending on the domain, getting such data could still be difficult and expensive. Having problem domain data results in weaker knowledge than original data.

\textbf{Non-Problem Domain (NPD) data} \cite{correia-silva_copycat_2018} is data sampled from the same type of content as the target model’s input, e.g. image data for image models and text data for text models. This data has the same syntactic type, potentially the same feature space, but a rather different marginal probability distribution. If there is any public data of the same modality as the original model, any attacker can use it and hence we can consider NPD data as the weakest knowledge.

\textbf{Artificial data} includes e.g. data produced by GANs \cite{kariyappa_maze_2021}, data sampled from standard probability distributions \cite{tramer_stealing_2016}, noise \cite{roberts_model_2019}, and data obtained as the result of optimisation of the input space without using any natural samples \cite{yuan_es_2022, gong_inversenet_2021}. Depending on the method used, artificial data can be more or less valuable than, e.g. NPD data. Hence, we can not unequivocally say how strong an attacker is with artificial data without knowing the properties of this data.

Papernot et al. experimented with stealing a model trained on the MNIST dataset \cite{papernot_practical_2017}. They performed an attack using a handcrafted digit dataset for model querying, assuming that the original data is not available. 
%
Correia-Silva et al. proposed an attack called "Copycat" which trains a substitute for a target CNN using NPD data \cite{correia-silva_copycat_2018}.
Orekondy et al. also used NPD data to train a substitute model (a "Knockoff net") for stealing CNNs \cite{orekondy_knockoff_2019}. 
In their results, a substitute model trained on the original data performs better than the one trained on NPD data.
Later, Zhang et al. explored how the attacker’s knowledge affects the attack performance if the attacker’s dataset only covers a few classes of the original dataset, or if there is only non-problem domain data available \cite{zhang_thief_2021}. 

Gong et al. proposed to leverage a model inversion attack in their SMA called InverseNet \cite{gong_inversenet_2021}. First, they trained a simple substitute model on data selected from public datasets. 
Based on this model, they selected samples with high confidence scores as starting point for a model inversion attack to obtain representative samples for each class. 
After being augmented, these samples are used to query the target model to train a final substitute model. 

Having data with a distribution similar to the original dataset can be crucial for the substitute model performance, especially for complex classification tasks. However, since it might be challenging to obtain such data, several works consider so-called "data-free" scenarios, where an attacker creates artificial data, assuming only little or no available natural data.
Kariyappa et al. \cite{kariyappa_maze_2021} proposed a model stealing attack called "MAZE", which uses a generative model that works similar to GANs, but learns to generate those samples on which attacker and target models disagree the most. 
They also considered a case in which an adversary has access to a small subset of original training data. In these settings, they trained a Wasserstein GAN to generate artificial samples. This attack performed better and required significantly fewer queries than the initial attack.
Yuan et al. proposed an attack called "ES Attack" \cite{yuan_es_2022}. It consists of two key steps: 
estimation of the parameters of the substitute model and data synthesising. The authors presented two methods for crafting artificial samples: the first uses an Auxiliary Classifier GAN for data generation, and the second operates directly in the input space.

Truong et al. launched a data-free model extraction (DFME) attack to steal CNNs \cite{truong_data-free_2021}. 
They trained a generator that produces samples in which the target and the substitute models disagree the most. 
Miura et al. introduced MEGEX -- an adaptation of the DFME attack for the case when confidence scores and gradient-based explanations are returned for each query \cite{miura_megex_2021}. 
Sanyal et al. introduced DFMS-HL, an SMA that requires no data and, in contrast to other data-free attacks \cite{kariyappa_maze_2021, truong_data-free_2021, miura_megex_2021} uses only top-1 labels for stealing CNNs \cite{sanyal_towards_2022}. From a (randomly) initialised GAN, they iteratively generate samples that are labelled by the target model, and are used for both training the substitute model and improving the GAN. 
The generator was trained using adversarial loss and class diversity losses, whereby the latter allows reaching an almost uniform distribution of generated samples across all classes. 
Xie et al. also used a GAN to produce samples for their attack called GAME \cite{xie_game_2022}. However, they assumed that (N)PD data is available and used it to train an auxiliary classifier GAN (AC-GAN). Through active learning they determined the most promising classes and then used the AC-GAN to generate samples from these classes to train a substitute model.

While attacking a classification model, an adversary might obtain different levels of detail from the target model, e.g., confidence scores for each class, or just top-1 labels. While confidence scores might contain important information for an attack, but are not always available. A few works proposed how to imitate them having only top-1 labels available. Wang et al. proposed the Black-Box Dissector, an SMA which operates with NPD data \cite{wang_black-box_2022}. 
They showed how to estimate the confidence scores of the target model by erasing parts of images (selected by, e.g. the Gradient-weighted Class Activation Mapping (Grad-CAM) method \cite{selvaraju_grad-cam_2017}) and aggregating predictions for them.  
Wang and Lin proposed another method of emulating class probabilities \cite{wang_enhance_2022}.
For each class, they first created a prototypical representation, and then set the probability of belonging to a certain class based on the distance to the corresponding representations.

Roberts et al. provided an SMA using only noise as input for querying \cite{roberts_model_2019}. They experimented with noise coming from different distributions: Uniform, Standard Normal, Standard Gumbel, Bernoulli, and Ising.
The authors claim that their attack is a parameter stealing attack; however, since they are not stealing model parameters directly but instead observe them by substitute model training, we classify this attack as behaviour stealing attack.

Mosafi et al. proposed an approach using a composite data generation method \cite{mosafi_stealing_2019}. They created a new dataset from a public one by superimposing two randomly selected images and used it to train a substitute model. The authors showed that a substitute model trained using the superimposed images -- even if only predicted labels are available from the model -- performs better than a model trained on regular data with confidence scores, i.e. a more detailed output.

\subsubsection{Number of Queries}
Another important aspect is the number of samples sent to the black-box model for labelling (i.e. the number of queries), since this is one of the most critical metrics in terms of the attack's efficiency -- and also a potential way for a defender to detect attacks. The strongest assumption in this case means having an attacker with no limits on the number of queries. In weaker settings, an adversary has a limited number of queries available and hence applies different techniques to reduce them. 
These techniques include, for instance, picking the most informative samples for querying, making samples more informative by crafting adversarial examples, or augmenting the attacker's dataset. 
Where available, we have gathered information on the number of queries (absolute as well as relative to the number of parameters of the model) in \Cref{tab:attacksPerformance-SMA}.

Three optimisation techniques have frequently been proposed to pick the most optimal samples for queries: active learning, reinforcement learning, and evolutionary algorithms.
Active learning is one of the most widely explored techniques for optimising the querying process. Tramèr et al. were the first to propose optimising queries \cite{tramer_stealing_2016}, suggesting two optimisation strategies: besides line-search (samples laying close to the decision boundary), they use adaptive retraining based on active learning. Later Reith et al. used adaptive retraining to steal SVRs \cite{reith_efficiently_2019}. 
Chandrasekaran et al. \cite{chandrasekaran_exploring_2020} showed how two active learning approaches, namely \textit{probably approximately correct} (PAC) and \textit{query synthesis} (QS), can be applied to steal DTs, RFs, LBMs, SVMs. 
Shi et al. \cite{shi_active_2018} used an active learning approach that reveals uncertain samples which are in turn used as queries to steal FNN (MLP). 
Pengcheng et al. \cite{pengcheng_query-efficient_2018} compared a non-optimised random selection strategy with two active learning methods: least confidence and margin-based. 
Pal et al. explored active learning strategies such as uncertainty, K-center and  DeepFool-based Active Learning (DFAL) to identify the most meaningful samples and, subsequently, use them for their "Activethief" attack \cite{pal_activethief_2020}. 
Several other works used active learning together with non-problem domain or artificial data for stealing image classifiers \cite{pal_framework_2019, wang_enhance_2022, xie_game_2022}.

Reinforcement learning for picking optimal samples was first utilised by Orekondy et. al. in their Knockoff attack \cite{orekondy_knockoff_2019}. Zhang et al. demonstrated that an attacker applying reinforcement learning and adversarial examples for the querying process performs better than having non-problem domain samples with no query strategy \cite{zhang_thief_2021}.
Barbalau et al. \cite{barbalau_black-box_2020} assumed that for an adversary without problem-domain data, it could be difficult to generate samples that are classified with high confidence by the target model. 
Hence, they applied an evolutionary algorithm to select from samples generated by a GAN those that will be predicted with high confidence.

Another way to optimise queries is to generate samples that help an adversary train a model with better performance. 
Adversarial examples were first ustilised by Papernot et al., who used Jacobian-Based Data Augmentation (JBDA) to generate new samples that are close to the decision boundary \cite{papernot_practical_2017}. Juuti et al. \cite{juuti_prada_2019} and Pengcheng et al. \cite{pengcheng_query-efficient_2018} crafted training samples using white-box adversarial example generation techniques, for instance the fast gradient sign method (FGSM, \cite{goodfellow_explaining_2015}). 
Yu et al. combined active learning and adversarial examples to reduce the number of queries in their attack FeatureFool \cite{yu_cloudleak_2020} which exploits benign and adversarial examples with different, but low target model confidence scores.
We note here that, in some papers, the authors used adversarial examples rather to reach a high transferability of the attack, instead of optimising the number of queries. However, this approach is promising for both goals. We also notice that adversarial crafting was mainly applied to original or problem-domain data, which might mean that this approach only works for an attacker with stronger data knowledge.

Data augmentation techniques enlarge the attacker's dataset while spending less queries. 
Shi et al. proposed to optimise the number of queries by using GAN-based data augmentation \cite{shi_generative_2018}. They first queried the target model with a small number of samples, then used the dataset thus obtained to train a GAN, which was in turn used to produce training data for the substitute model training.
We distinguish attack approaches that use generative models to increase the quality of the attacker's data, mentioned in \Cref{sec:sma-data}, from attacks that aim to augment the attacker's dataset.
In the first case, the number of queries is usually much bigger (see \Cref{tab:attacksPerformance-SMA}), and an attacker does not optimise the queries, as in the second case.

Besides query optimisation, a few other strategies for attack improvement were proposed. Joshi et al. suggest to use a gradient-driven adaptive learning rate (GDLR) to make the substitute model learning process more efficient \cite{joshi_gdalr_2019}. 
Aivodji et al. \cite{aivodji_model_2020} showed how MLPs can be stolen using counterfactual explanations (CFEs) in addition to the regular labels.
Counterfactual explanations are very similar to adversarial examples, whereby their intent is not to deceive the model, but rather to explain it \cite{molnar_interpretable_2019}.
These samples, together with initially queried ones, are used for substitute model training. 
Wang et al. argue that using CFEs and regular queries to train a substitute model leads to shifting the decision boundary far away from the one of the target model \cite{wang_dualcf_2022}, since regular queries usually lay far from the decision boundary, while CFEs are close to it. Thus, if a model tries to separate them, the decision boundary shifts towards the regular samples. 
To overcome this issue, the authors introduced DualCF, an attack that additionally uses CFEs of CFEs (named CCFEs), thus creating samples close to both sides of the decision boundary. They also theoretically proved that a single couple of CFE and CCFE is enough to extract a linear model with 100\% fidelity. 

\subsection{Substitute Model Training Attacks against Specific Model Types and Domain-specific Models}
\label{sec:substitute:domainspecific}
Substitute model attacks can be widely used for different models and data domains. Below we provide an overview of works that explore these attacks in very specific or highly-focused settings. Since most of the research is dedicated to CNNs and classification, attacks and defences on other types of models are less explored and remain an open topic.

Takemura et al. \cite{takemura_model_2020} explored attacks against LSTM for both classification- and regression tasks. They trained an LSTM as well as RNN with lower complexity as a substitute model, and showed that the substitute LSTM performs better than the RNN. 
Krishna et al. \cite{krishna_thieves_2020} explored model stealing attacks against BERT-based models \cite{devlin_bert_2019}, which are commonly used in natural language processing (NLP). They showed how all of transfer learning, a mismatch between target- and substitute architectures, and the attacker's data source affect the substitute model performance. 
He et al. also explored SMAs against BERT-based models and evaluated their transferability scores \cite{he_model_2021}. They also showed that the accuracy of the attack remains high even if there is a mismatch in the architectures. 

Behzadan and Hsu \cite{behzadan_adversarial_2019} investigated a model extraction attack against deep reinforcement learning. They utilised a technique called "Deep Q-Learning from Demonstrations" to develop two attacks that learn an adversarial policy - i.e. an imitation of the target policy. 
They first tried to predict the training algorithm family based on an action sequence, using an RNN. Then, the authors utilised imitation learning \cite{hussein_imitation_2017} to train a substitute DRL model. 

Szyller et al. target image transformation models \cite{szyller_good_2021}, and stole the functionality of GANs for neural style transfer and super-resolution tasks by training a substitute model on image pairs obtained from queries. 
Hu and Pang trained a substitute GAN on images generated by the target GAN for two scenarios: high-fidelity- and high-accuracy extraction \cite{hu_stealing_2021}. The main distinction is that for high-accuracy extraction, they applied an additional step of subsampling high-quality samples by using the discriminator of the target GAN.

Liu et al. launched an SMA (called StolenEncoder) against encoders trained in self-supervised settings using contrastive learning \cite{liu_stolenencoder_2022}. They queried the target encoder with images to obtain the original embeddings and trained a substitute encoder so that its embeddings coincided with the original. 
To reach better performance while using fewer queries, they augmented images and used embeddings of corresponding non-augmented images as ground truth. 
Sha et al. considered the same settings and proposed another attack that uses contrastive learning, called ContSteal \cite{sha_cant_2022}. They defined a loss function that minimises the difference between target- and substitute embeddings of the same image and maximises it for different images. 
Dziedzic et al. explored three attack scenarios against encoders \cite{dziedzic_difficulty_2022}. In some settings, using the original projection head is beneficial for the downstream classification accuracy.

Several works explore model stealing attacks against graph neural networks (GNNs). 
Since graphs are a collection of nodes and edges, they generally contain more degrees of freedom and using random data as attacker's data is less effective. 
DeFazio et al. were the first to introduce a GNN model stealing method \cite{defazio_adversarial_2020}. They considered a node classification problem and proposed an attack that allows to steal a 2-layer GNN, if knowing a subset of original training data and having access to a 2-hop subgraph of the original graph. 
Wu et al. \cite{wu_model_2022} also attack GNNs for node classification problems. However, the authors explored different settings regarding the attacker's knowledge: the adversary may know node attributes, the graph structure and/or have access to shadow (auxiliary) data. 
He et al. \cite{he_stealing_2021} also explored these settings while attacking GNNs; however, in contrast to previous works, they considered link stealing attacks that aim to reveal if two nodes are connected. 
Shen et al. considered SMAs against inductive GNNs, i.e. GNNs that can infer previously unseen unlabelled data \cite{shen_model_2022}. They studied two attack types: with and without knowledge of the target GNN structure. For each of them, they considered three different attackers depending on the available output information. 

Teitelman et al. proposed an attack that steals the functionality of a microchip \cite{teitelman_stealing_2020}.
They introduced an architecture called Deep Neural Tree, which is a combination of a neural network and a decision tree. This model can learn to distinguish different tasks of the chip and provide a certain level of explainability thanks to the tree-like architecture. 

Yan et al. proposed a dual‐task model extraction attack (DTMEA) \cite{yan_towards_2022} for stealing a model that returns both confidence scores and output explanations by training a multi-task CNN with two classification heads: one solves the classification task, and another learns to imitate explanations. Although stealing explanations is not the primary goal of model stealing, the multi-task substitute model reached higher accuracy than a substitute trained only on confidence scores.

Ali and Eshete \cite{ali_best-effort_2020} explored SMA against malware classifiers for Windows Portable Executables (PEs). In particular, they use different data representations for the target model (features extracted from bytes) and a substitute model (images based on bytes-to-pixels mapping).
They also experimented with mismatching architectures for target and substitute models and concluded that a similar architecture is not the best choice. In their experiments, using a pre-trained Inception-V3 as a substitute model resulted in higher fidelity than using a custom MalConv model (CNN for Malware detection) that corresponds to the target model architecture. 
We speculate, that one of the reasons could be that the size of the substitute training set was only 40\% of the size of the target training set; hence, transfer learning could play a crucial role.
Yue et al. explored SMA against a sequential recommender system (SRS), aiming to open the black-box target model for performing further profile pollution and data poisoning attacks \cite{yue_black-box_2021}. 

Aarts et al. considered a different goal for creating substitute models \cite{aarts_leveraging_2021}. Instead of creating a high-performing substitute that completely emulates a target model, they proposed to train a substitute to a certain level of effectiveness and then, depending on its confidence scores, use either the substitute or the target model to obtain predictions. 
This scenario is feasible if an attacker steals the model to launch a competitive API since, in this case, even by delegating some of the queries to the target model and paying for them, the attacker can turn a profit. 

\subsection{Meta-model Training Attacks} 
\label{sec:meta-model}
Oh et al. \cite{oh_towards_2018} proposed the meta-model attack (MMA) -- the first and so far only query-based attack that can reveal information about the target model architecture. They trained a meta-model that, for a given model, predicts details about the target model structure, training setup, and the amount of training data. 
As dataset for the meta-model, they used a set of candidate CNNs which vary in architecture parameters (type of activation functions, the number of convolutional and fully-connected layers, etc.), optimisation parameters (type of algorithm, batch size), and data parameters (data split, data size).
Then the meta-model was trained to represent the correlation between hyperparameters of a model and its performance on specific test samples. Those samples were subsequently used to reveal the hyperparameters of the target model. One peculiarity of the attack is that to successfully steal a hyperparameter, this hyperparameter should be influential for the target model and its value should appear in the training set. For instance, to steal the number of convolutional layers, an adversary has to be sure that the target model is indeed a convolutional neural network and that in the training set of the meta-model there is a model with the same number of convolutional layers.
The attack requires significant computational power- and time resources. For instance, to perform hyperparameter stealing on MNIST classifiers, the authors created 10,000 candidate CNNs which took 40 days of training on a GPU. On average, the attack predicted the correct hyperparameter value in 80.1\% of the cases, whereas the average chance to guess is 34.9\%.
Given that the meta-model attack steals the hyperparameters of the model, an adversary needs an additional parameter-stealing attack to obtain a model that approximates the target model behaviour. For the same reason, this attack cannot be compared with other query-based attacks in terms of effectiveness and efficiency.

\subsection{Comparison of Query-Based Attack Performance}
In this section, we compare the performance of query-based model stealing attacks in terms of effectiveness and efficiency of the attack, with the help of \Cref{tab:attacksPerformance-non-SMA,tab:attacksPerformance-SMA}, whereby the latter covers substitute model attacks and the former other query-based attacks.
For each of these attacks, the tables provide the type of models they have been applied to -- for both the target model (TM) to be stolen and the model type chosen by the attacker, i.e. the attacker model (AM)) -- and the data modality considered in the evaluation.
\newcolumntype{L}[1]{>{\raggedright\let\newline\\\arraybackslash\hspace{0pt}}m{#1}}
\newcommand\scalemath[2]{\scalebox{#1}{\mbox{\ensuremath{\displaystyle #2}}}}
\begin{table}[]
\centering
\scriptsize
\setlength{\tabcolsep}{1pt}
\setlength{\arraycolsep}{1pt}
\renewcommand{\arraystretch}{1.05}
\caption{Performance and other characteristics of query-based attacks exluding SMAs. N/A indicates that the authors did not provide the information}
\label{tab:attacksPerformance-non-SMA}
\resizebox{\linewidth}{!}{%
\begin{tabular}{
|L{0.182\linewidth}
|L{0.084\linewidth}
|L{0.082\linewidth}
|L{0.060\linewidth}
|L{0.071\linewidth}
|L{0.068\linewidth}
|L{0.100\linewidth}
|L{0.090\linewidth}
|L{0.210\linewidth}
|
}
\hline
\multicolumn{1}{|c|}{\multirow{2}{*}{\textbf{Attack}}} &
  \multicolumn{2}{c|}{\textbf{Model}} &
  \multicolumn{1}{c|}{\textbf{Data}} &
  \multicolumn{2}{c|}{\textbf{TM Properties}} &
  \multicolumn{3}{c|}{\textbf{Stealing Performance}} \\ \cline{2-3} \cline{5-9} 
\multicolumn{1}{|c|}{} &
  \multicolumn{1}{l|}{\textit{Target (TM)}} &
  \textit{Attack (AM)} &
  \multicolumn{1}{c|}{\textbf{Modality}} &
  \multicolumn{1}{l|}{\textit{Effectiveness}} &
  \textit{Parameters} &
  \multicolumn{1}{l|}{\textit{Query budget}} &
  \multicolumn{1}{l|}{\textit{Efficiency Score}} &
  \textit{Stealing Effectiveness} \\ \hline
\multirow{2}{0.99\linewidth}{WFA \tiny{\cite{lowd_adversarial_2005}}} &
  \tiny{NB} &
  \tiny{Same as target} &
  \multirow{2}{0.99\linewidth}{\tiny{Tabular}} &
  \tiny{\nonap} &
  \tiny{23k; 1k; 1k} &
  \tiny{261k; 25k; 23k} &
  \tiny{11; 25; 23} &
  \tiny{Exact extraction} \\ \cline{2-3} \cline{5-9} 
 &
  \tiny{MaxEnt} &
  \tiny{Same as target} &
   &
  \tiny{\nonap} &
  \tiny{23k; 1k; 1k} &
  \tiny{119k; 10k; 9k} &
  \tiny{5; 10; 9} &
  \tiny{Exact extraction} \\ \hline
WFA \tiny{\cite{tramer_stealing_2016}} &
  \tiny{LogReg} &
  \tiny{Same as target} &
  \tiny{Tabular} &
  \tiny{\nonap} &
  \tiny{d} &
  \tiny{50d} &
  \tiny{50} &
  \tiny{Exact extraction} \\ \hline
WFA \tiny{\cite{reith_efficiently_2019}} &
  \tiny{SVM-lin} &
  \tiny{Same as target} &
  \tiny{Tabular} &
  \tiny{\nonap} &
  \tiny{d} &
  \tiny{17d} &
  \tiny{17} &
  \tiny{Exact extraction} \\ \hline
ESA \tiny{\cite{tramer_stealing_2016}} &
  \tiny{(M)LogReg, NN} &
  \tiny{Same as target} &
  \tiny{Tabular} &
  \tiny{\unknown} &
  \tiny{d} &
  \tiny{d; |classes| $\scalemath{0.7}{\times}$ d; 4d} &
  \tiny{1; |classes|; 4} &
  \tiny{100\% fid; 100\% fid; 99.99\% fid} \\ \hline
ESA \tiny{\cite{reith_efficiently_2019}} &
  \tiny{SVR-lin/quad} &
  \tiny{Same as target} &
  \tiny{Tabular} &
  \tiny{\unknown} &
  \tiny{d} &
  \tiny{d;$\frac{\text{\tiny{1}}}{\text{\tiny{2}}}\text{\tiny{d}}^{\text{\tiny{2}}} + \frac{\text{\tiny{3}}}{\text{\tiny{2}}}$d+1} &
  \tiny{1; $\frac{\text{\tiny{1}}}{\text{\tiny{2}}}$d+$\frac{\text{\tiny{3}}}{\text{\tiny{2}}}$} &
  \tiny{100\% fid} \\ \hline
ESA \tiny{(dupl. quer.)} \tiny{\cite{yan_monitoring-based_2021}} &
  \tiny{LogReg, NN} &
  \tiny{Same as target} &
  \tiny{Tabular} &
  \tiny{\unknown} &
  \tiny{d} &
  \tiny{|duplications|$\scalemath{0.7}{\times}$d} &
  \tiny{|duplications|} &
  \tiny{~88-100\% fid; ~98-100\% acc} \\ \hline
\multirow{2}{0.99\linewidth}{PFA \tiny{(compl. quer.; incompl. quer.)} \tiny{\cite{tramer_stealing_2016}}} &
  \tiny{DT} &
  \multirow{2}{0.99\linewidth}{\tiny{Same as target}} &
  \multirow{2}{0.99\linewidth}{\tiny{Tabular}} &
  \multirow{2}{0.99\linewidth}{\tiny{\unknown}} &
  \tiny{26-318\tablefootnote{\label{non-param}Since DTs and RTs are non-parametric, we use the number of leaves in a tree.}} &
  \tiny{1.7k-101k; 1.1k-30k} &
  \tiny{19-318; 17-100} &
  \tiny{86.4-100\% fid; 99.65-100\% fid} \\ \cline{2-2} \cline{6-9} 
 &
  \tiny{RT} &
   &
   &
   &
  \tiny{49-155\cref{non-param}} &
  \tiny{6k-32k; 1.8k-7.4k} &
  \tiny{122-206; 36-48} &
  \tiny{100\% fid; 100\% fid} \\ \hline
Recovery \tiny{\cite{jagielski_high_2020}} &
  \tiny{ReLU NN} &
  \tiny{Same as target} &
  \tiny{Image} &
  \tiny{94.3-97.7\%} &
  \tiny{12,5k - 100k} &
  \tiny{$\scalemath{0.9}{2^{\scalemath{0.8}{17.2}}}$ - $\scalemath{0.9}{2^{\scalemath{0.8}{20.2}}}$} &
  \tiny{12} &
  \tiny{99.98-100\% fid} \\ \hline
Recovery \tiny{\cite{rolnick_reverse-engineering_2020}} &
  \tiny{ReLU DNN} &
  \tiny{Same as target (isomorphic)} &
  \tiny{Image, Tabular} &
  \tiny{\unknown} &
  \tiny{\unknown} &
  \tiny{\unknown} &
  \tiny{250-390 (est)} &
  \tiny{\unknown} \\ \hline
Recovery \tiny{\cite{carlini_cryptanalytic_2020}} &
  \tiny{ReLU DNN} &
  \tiny{Same as target} &
  \tiny{\unknown} &
  \tiny{\unknown} &
  \tiny{210 – 100k} &
  \tiny{$\scalemath{0.9}{2^{\scalemath{0.8}{16}}}$ - $\scalemath{0.9}{2^{\scalemath{0.8}{21.5}}}$} &
  \tiny{30 - 312} &
  \tiny{100\% fid} \\ \hline
\end{tabular}%
}
\end{table}

Further, the tables provide details on the performance of the attacks.
We report the effectiveness of the target model as reference (by default in \% accuracy) as well as the number of parameters to be stolen, i.e. the number of learned parameters of the target model.
Also, the reported number of queries is shown; in several works, multiple settings are evaluated, e.g. an attack with a small, medium and large number of queries (and corresponding other effectiveness measures); these are also shown in the table.
If both the number of parameters and queries are given, we can compute the relation of queries per parameter as \textit{efficiency score}. Finally, the tables detail the reported stealing effectiveness, i.e. the accuracy, fidelity or transferability (cf. \Cref{sec:performance_objectives}), or other scores, e.g. AUC.
 
There are two categories of data not provided in the table. (1) The effectiveness score of the target model for WFAs (denoted as grey cells). As WFAs produce an exact copy of the target model, the accuracy of the target model equals the accuracy of the stolen model. Hence, the relative accuracy is always $100\%$, and we do not need to know the effectiveness of the target model. 
%
\begin{table}[]
\centering
\scriptsize
\setlength{\tabcolsep}{1pt}
\setlength{\arraycolsep}{1pt}
\renewcommand{\arraystretch}{1.05}
\caption{Performance and other characteristics of SMAs. N/A indicates that the authors did not provide the information}
\label{tab:attacksPerformance-SMA}
\resizebox{\linewidth}{!}{%
\begin{tabular}{
|L{0.182\linewidth}
|L{0.084\linewidth}
|L{0.082\linewidth}
|L{0.060\linewidth}
|L{0.071\linewidth}
|L{0.068\linewidth}
|L{0.100\linewidth}
|L{0.090\linewidth}
|L{0.210\linewidth}
|
}
\hline
\multicolumn{1}{|c|}{\multirow{2}{*}{\textbf{Attack}}} &
  \multicolumn{2}{c|}{\textbf{Model}} &
  \multicolumn{1}{c|}{\textbf{Data}} &
  \multicolumn{2}{c|}{\textbf{TM Properties}} &
  \multicolumn{3}{c|}{\textbf{Stealing Performance}} \\ \cline{2-3} \cline{5-9} 
\multicolumn{1}{|c|}{} &
  \multicolumn{1}{l|}{\textit{Target (TM)}} &
  \textit{Attack (AM)} &
  \multicolumn{1}{c|}{\textbf{Modality}} &
  \multicolumn{1}{l|}{\textit{Effectiveness}} &
  \textit{Parameters} &
  \multicolumn{1}{l|}{\textit{Query budget}} &
  \multicolumn{1}{l|}{\textit{Efficiency Score}} &
  \textit{Stealing Effectiveness} \\ \hline
SMA-\tiny{*/NN (retraining)} \tiny{\cite{tramer_stealing_2016}} &
  \tiny{MLogReg; NN; SVM-RBF} &
  \tiny{Same as target} &
  \tiny{Tabular} &
  \tiny{\unknown} &
  \tiny{d} &
  \tiny{100d$\scalemath{0.7}{\times}$|classes|; 100d; 10d-100d} &
  \tiny{ 100$\scalemath{0.7}{\times}$|classes|; 100; 10-100} &
  \tiny{98.24-100\% fid} \\ \hline
SMA-* \tiny{(retraining)} \tiny{\cite{reith_efficiently_2019}} &
  \tiny{SVM-lin/RBF, SVR-RBF} &
  \tiny{Same as target} &
  \tiny{Tabular} &
  \tiny{\unknown} &
  \tiny{d} &
  \tiny{d; 20d;} \tiny{d-40d} &
  \tiny{1; 20;} \tiny{1-40} &
  \tiny{99-100\% fid} \\ \hline
SMA-\tiny{NN} \tiny{\cite{papernot_practical_2017}} &
  \tiny{CNNs, LogReg, SVM, DT, kNN} &
  \tiny{LogReg, CNNs } &
  \tiny{Image} &
  \tiny{92-94.97\%} &
  \tiny{60k (est)} &
  \tiny{6.4k} &
  \tiny{9 (est)} &
  \tiny{61-89\% fid, 96-97\% tr} \\ \hline
SMA-\tiny{NN} \tiny{\cite{yi_shi_how_2017}} &
  \tiny{NN, SVM, NB} &
  \tiny{SVM, NB, NN} &
  \tiny{Text} &
  \tiny{85.56-96.51\%} &
  \tiny{\unknown} &
  \tiny{859} &
  \tiny{\unknown} &
  \tiny{97.44 - 97.9\% fid} \\ \hline
SMA-\tiny{CNN (Copycat)} \tiny{\cite{correia-silva_copycat_2018}} &
  \tiny{VGG-16} &
  \tiny{Same as target} &
  \tiny{Image} &
  \tiny{88.7-95.8\%} &
  \tiny{138m (est)} &
  \tiny{3m} &
  \tiny{2$\scalemath{0.7}{\times} \scalemath{0.9}{10^{\scalemath{0.8}{-2}}}$ (est)} &
  \tiny{93.7-98.6\% rel acc} \\ \hline
\multirow{3}{0.99\linewidth}{SMA-\tiny{CNN;RNN (Activethief)} \tiny{\cite{pal_activethief_2020}}} &
  \tiny{CNN} &
  \multirow{3}{0.99\linewidth}{\tiny{Same as target}} &
  \tiny{Image} &
  \multirow{3}{0.99\linewidth}{\tiny{\unknown}} &
  \multirow{3}{0.99\linewidth}{\tiny{\unknown}} &
  \tiny{10k-120k} &
  \multirow{3}{0.99\linewidth}{\tiny{\unknown}} &
  \tiny{64.2\%-95.8\% fid (10k)84.99-98.54 fid (120k)} \\ \cline{2-2} \cline{4-4} \cline{7-7} \cline{9-9} 
 &
  \tiny{CNN \cite{kim_convolutional_2014}} &
   &
  \multirow{2}{0.99\linewidth}{\tiny{Text}} &
   &
   &
  \tiny{10k-89k} &
   &
  \tiny{75.87-77.69 fid (10k)86.21-90.07 fid (89k)} \\ \cline{2-2} \cline{7-7} \cline{9-9} 
 &
  \tiny{RNN (GRU)} &
   &
   &
   &
   &
  \tiny{89k} &
   &
  \tiny{89.12-93.01 fid (89k)} \\ \hline
SMA-\tiny{CNN} \tiny{\cite{juuti_prada_2019}} &
  \tiny{CNN, VGG-16} &
  \tiny{Same as target} &
  \tiny{Image} &
  \tiny{95-98\%} &
  \tiny{CNN: 486k} &
  \tiny{102k; 6.4k} &
  \tiny{2.1$\scalemath{0.7}{\times} \scalemath{0.9}{10^{\scalemath{0.8}{-1}}}$; 1.3$\scalemath{0.7}{\times} \scalemath{0.9}{10^{\scalemath{0.8}{-2}}}$} &
  \tiny{97.9\% fid; 39.3 \% tr} \\ \hline
SMA-\tiny{CNN (Knockoff)} \tiny{\cite{orekondy_knockoff_2019}} &
  \tiny{ResNet34} &
  \tiny{Same as target*\tablefootnote{Same as target* means that we report results for the same architecture but the authors also provide results for other substitute architectures.}} &
  \tiny{Image} &
  \tiny{78.8\%} &
  \tiny{21M (est)} &
  \tiny{60k} &
  \tiny{2.9$\scalemath{0.7}{\times} \scalemath{0.9}{10^{\scalemath{0.8}{-3}}}$ (est)} &
  \tiny{76.2\% acc (97\% rel acc)} \\ \hline
SMA-\tiny{CNN (Knockoff)} \tiny{\cite{atli_extraction_2020}} &
  \tiny{ResNet34} &
  \tiny{Same as target} &
  \tiny{Image} &
  \tiny{71.1-98.1\%} &
  \tiny{21M (est)} &
  \tiny{1.2m (est)} &
  \tiny{6$\scalemath{0.7}{\times} \scalemath{0.9}{10^{\scalemath{0.8}{-2}}}$ (est)} &
  \tiny{53.5-94.8\% acc (75-97\% rel acc)} \\ \hline
SMA-\tiny{NN (DS)} \tiny{\cite{shi_generative_2018}} &
  \tiny{\unknown} &
  \tiny{NN} &
  \tiny{Tabular} &
  \tiny{\unknown} &
  \tiny{\unknown} &
  \tiny{100} &
  \tiny{\unknown} &
  \tiny{69.2-72.6\% fid} \\ \hline
SMA-\tiny{NN (AL)} \tiny{\cite{shi_active_2018}} &
  \tiny{\unknown} &
  \tiny{NN} &
  \tiny{Tabular} &
  \tiny{\unknown} &
  \tiny{\unknown} &
  \tiny{1k} &
  \tiny{\unknown} &
  \tiny{80.77\% fid} \\ \hline
\multirow{2}{0.99\linewidth}{SMA-\tiny{CNN (DS + AL)} \tiny{\cite{pengcheng_query-efficient_2018}}} &
  \tiny{CNN} &
  \tiny{CNN (simple)} &
  \multirow{2}{0.99\linewidth}{\tiny{Image}} &
  \tiny{99.24\%} &
  \multirow{2}{0.99\linewidth}{\tiny{\unknown}} &
  \multirow{2}{0.99\linewidth}{\tiny{100-25.6k}} &
  \multirow{2}{0.99\linewidth}{\tiny{\unknown}} &
  \tiny{47.64-94.19\% fid} \\ \cline{2-3} \cline{5-5} \cline{9-9} 
 &
  \tiny{ResNet} &
  \tiny{VGG-16} &
   &
  \tiny{91\%} &
   &
   &
   &
  \tiny{53.61-79.75\% fid} \\ \hline
\multirow{2}{0.99\linewidth}{SMA-* \tiny{(AL)} \tiny{\cite{chandrasekaran_exploring_2020}}} &
  \tiny{SVM-kernel} &
  \multirow{2}{0.99\linewidth}{\tiny{Same as target}} &
  \multirow{2}{0.99\linewidth}{\tiny{Tabular}} &
  \tiny{\unknown} &
  \multirow{2}{0.99\linewidth}{\tiny{\unknown}} &
  \tiny{48-1k} &
  \multirow{2}{0.99\linewidth}{\tiny{\unknown}} &
  \tiny{94.5-98.2\% acc} \\ \cline{2-2} \cline{5-5} \cline{7-7} \cline{9-9} 
 &
  \tiny{DT} &
   &
   &
  \tiny{52.1-86.8\%} &
   &
  \tiny{361-244k} &
   &
  \tiny{73.1-89.4\% acc} \\ \hline
\multirow{2}{0.99\linewidth}{SMA-\tiny{CNN (PD, AL)} \tiny{\cite{pal_framework_2019}}} &
  \multirow{2}{0.99\linewidth}{\tiny{CNNs}} &
  \multirow{2}{0.99\linewidth}{\tiny{CNNs}} &
  \tiny{Image} &
  \multirow{2}{0.99\linewidth}{\tiny{\unknown}} &
  \multirow{2}{0.99\linewidth}{\tiny{\unknown}} &
  \tiny{10k; 30k; 100k} &
  \multirow{2}{0.99\linewidth}{\tiny{\unknown}} &
  \tiny{64.2-95.8\%; 78.36-98.18\%; 81.57-98.81\% fid} \\ \cline{4-4} \cline{7-7} \cline{9-9} 
 &
   &
   &
  \tiny{Text} &
   &
   &
  \tiny{10k; 30k; 89k} &
   &
  \tiny{58.6-77.67\%; 71.8-87.04\%; 77.8-90.07\% fid} \\ \hline
SMA-\tiny{CNN (DS)} \tiny{\cite{mosafi_stealing_2019}} &
  \tiny{CNN} &
  \tiny{VGG-16} &
  \tiny{Image} &
  \tiny{90.48\%} &
  \tiny{\unknown} &
  \tiny{\unknown} &
  \tiny{\unknown} &
  \tiny{89.59\% acc} \\ \hline
SMA-\tiny{CNN (DS)} \tiny{\cite{yuan_es_2022}} &
  \tiny{LeNet5, ResNet18,34} &
  \tiny{LeNet5, ResNet18,34} &
  \tiny{Image} &
  \tiny{91.12-99.10\%} &
  \tiny{60k-60m (est)} &
  \tiny{\unknown} &
  \tiny{\unknown} &
  \tiny{80.79-93.97\% acc (88.66-94.82\% rel acc); 92.14-100\% tr} \\ \hline
\multirow{2}{0.99\linewidth}{SMA-\tiny{CNN (membership; gradients)} \tiny{\cite{milli_model_2019}}} &
  \tiny{MLogReg, ReLU-NN, CNN} &
  \multirow{2}{0.99\linewidth}{\tiny{Same as target, permuted}} &
  \tiny{Image: MNIST} &
  \tiny{93-99\% (est)} &
  \tiny{\unknown} &
  \tiny{CNN: 1k;10; MLogReg: 784;1; ReLU-NN: 10k;100} &
  \tiny{\unknown} &
  \tiny{93-99\% acc (est) } \\ \cline{2-2} \cline{4-9} 
 &
  \tiny{CNN, VGG-11, ResNet18} &
   &
  \tiny{Image: CIFAR10} &
  \tiny{75-90\% (est)} &
  \tiny{11m-15m (est; excl. CNN)} &
  \tiny{CNN: 10k;100; VGG/ResNet: 10k;1k} &
  \tiny{$\scalemath{0.9}{10^{\scalemath{0.8}{-5}}}$-$\scalemath{0.9}{10^{\scalemath{0.8}{-3}}}$ (excl. CNN)} &
  \tiny{75-88\% acc (est)} \\ \hline
SMA-\tiny{CNN} \tiny{\cite{kariyappa_maze_2021}} &
  \tiny{LeNet, ResNet20} &
  \tiny{WideResNet22} &
  \tiny{Image} &
  \tiny{91.04-97.43\%} &
  \tiny{\unknown} &
  \tiny{5M-30M} &
  \tiny{\unknown} &
  \tiny{82.9-94.32\% acc (91-99\% rel acc)} \\ \hline
SMA-\tiny{*/CNN (noisy)} \tiny{\cite{roberts_model_2019}} &
  \tiny{CNN} &
  \tiny{Same as target} &
  \tiny{Image} &
  \tiny{88.62-99.03\%} &
  \tiny{\unknown} &
  \tiny{600k} &
  \tiny{\unknown} &
  \tiny{10.47-95.93\% acc (11.81-96.87\% rel acc)} \\ \hline
 SMA-\tiny{CNN (DS)} \tiny{\cite{barbalau_black-box_2020}} &
  \tiny{AlexNet} &
  \tiny{half-AlexNet} &
  \tiny{Image} &
  \tiny{82.5\%} &
  \tiny{62M (est)} &
  \tiny{\unknown} &
  \tiny{\unknown} &
  \tiny{79.0\% acc} \\ \hline
SMA-\tiny{CNN} \tiny{\cite{ali_best-effort_2020}} &
  \tiny{CNN} &
  \tiny{Inception} &
  \tiny{Image} &
  \tiny{93\%} &
  \tiny{\unknown} &
  \tiny{16k} &
  \tiny{\unknown} &
  \tiny{88.65\% fid} \\ \hline
SMA-\tiny{CNN (FeatureFool)} \tiny{\cite{yu_cloudleak_2020}} &
  \tiny{\unknown} &
  \tiny{VGG-19-DeepID} &
  \tiny{Image} &
  \tiny{77.93\%} &
  \tiny{\unknown} &
  \tiny{2.15k} &
  \tiny{\unknown} &
  \tiny{76.05\% acc (97.63\% rel acc)} \\ \hline
SMA-\tiny{CNN (InverseNet)} \tiny{\cite{gong_inversenet_2021}} &
  \tiny{CNN} &
  \tiny{Same as target*} &
  \tiny{Image} &
  \tiny{\unknown} &
  \tiny{\unknown} &
  \tiny{30k} &
  \tiny{\unknown} &
  \tiny{95.88\% fid} \\ \hline
SMA-\tiny{CNN (DFME)} \tiny{\cite{truong_data-free_2021}} &
  \tiny{ResNet34} &
  \tiny{ResNet18} &
  \tiny{Image} &
  \tiny{95.5\%} &
  \tiny{21m (est)} &
  \tiny{20m} &
  \tiny{0.95 (est)} &
  \tiny{88.1\% acc (92\% rel acc)} \\ \hline
SMA-\tiny{CNN (MEGEX)} \tiny{\cite{miura_megex_2021}} &
  \tiny{ResNet34} &
  \tiny{ResNet18} &
  \tiny{Image} &
  \tiny{95.5\%} &
  \tiny{21m (est)} &
  \tiny{20m} &
  \tiny{0.95 (est)} &
  \tiny{92.3\% acc (97\% rel acc)} \\ \hline
SMA-\tiny{CNN} (DS) \tiny{\cite{sanyal_towards_2022}} &
  \tiny{ResNet34} &
  \tiny{ResNet18} &
  \tiny{Image} &
  \tiny{95.5\%} &
  \tiny{21m (est)} &
  \tiny{8m} &
  \tiny{0.38 (est)} &
  \tiny{93.96\% acc} \\ \hline
SMA-\tiny{CNN (NPD, Rl)} \tiny{\cite{zhang_thief_2021}} &
  \tiny{\unknown} &
  \tiny{\unknown} &
  \tiny{Image} &
  \tiny{\unknown} &
  \tiny{\unknown} &
  \tiny{10k} &
  \tiny{\unknown} &
  \tiny{75.4\% acc} \\ \hline
SMA-\tiny{CNN} \tiny{\cite{aarts_leveraging_2021}} &
  \tiny{MobileNetV2} &
  \tiny{Same as target} &
  \tiny{Image} &
  \tiny{\unknown} &
  \tiny{3m (est)} &
  \tiny{131k} &
  \tiny{4.4$\scalemath{0.7}{\times}\scalemath{0.9}{10^{\scalemath{0.8}{-2}}}$ (est)} &
  \tiny{100\% fid (est)} \\ \hline
SMA-\tiny{CNN} \tiny{\cite{wang_enhance_2022}} &
  \tiny{ResNet34} &
  \tiny{Same as target} &
  \tiny{Image} &
  \tiny{\unknown} &
  \tiny{21m (est)} &
  \tiny{30k} &
  \tiny{1.4$\scalemath{0.7}{\times} \scalemath{0.9}{10^{\scalemath{0.8}{-3}}}$ (est)} &
  \tiny{80.90\% acc} \\ \hline
SMA-\tiny{CNN} (Black-box Dissector) \tiny{\cite{wang_black-box_2022}} &
  \tiny{ResNet34} &
  \tiny{Same as target*} &
  \tiny{Image} &
  \tiny{91.56\%} &
  \tiny{21m (est)} &
  \tiny{30k} &
  \tiny{1.4$\scalemath{0.7}{\times} \scalemath{0.9}{10^{\scalemath{0.8}{-3}}}$  (est)} &
  \tiny{80.47\% acc, 82.14\% fid, 76.63\% tr} \\ \hline
SMA-\tiny{CNN} \tiny{\cite{yan_towards_2022}} &
  \tiny{ResNet50} &
  \tiny{CNN} &
  \tiny{Image} &
  \tiny{92.03\%} &
  \tiny{24m (est)} &
  \tiny{50k} &
  \tiny{2$\scalemath{0.7}{\times} \scalemath{0.9}{10^{\scalemath{0.8}{-3}}}$ (est)} &
  \tiny{85\% acc (est)} \\ \hline
SMA-\tiny{CNN} (GAME) \tiny{\cite{xie_game_2022}} &
  \tiny{AlexNet} &
  \tiny{Half-AlexNet} &
  \tiny{Image} &
  \tiny{98.29\%} &
  \tiny{62m (est)} &
  \tiny{\unknown} &
  \tiny{\unknown} &
  \tiny{75.88\% acc (77\% rel acc), 76.74\% fid} \\ \hline
SMA-* \tiny{(CFEs)} \tiny{\cite{aivodji_model_2020}} &
  \tiny{NN} &
  \tiny{Same as target} &
  \tiny{Tabular} &
  \tiny{84.7\%} &
  \tiny{\unknown} &
  \tiny{1k} &
  \tiny{\unknown} &
  \tiny{94.89 \% fid; 83.97 \% acc} \\ \hline
SMA-* \tiny{(DualCF)} \tiny{\cite{wang_dualcf_2022}} &
  \tiny{MLP} &
  \tiny{Same as target} &
  \tiny{Tabular} &
  \tiny{\unknown} &
  \tiny{\unknown} &
  \tiny{329} &
  \tiny{\unknown} &
  \tiny{99\% fid (est)} \\ \hline
SMA-\tiny{RNN} \tiny{\cite{krishna_thieves_2020}} &
  \tiny{BERT} &
  \tiny{BERT, XLNet} &
  \tiny{Text} &
  \tiny{76.1-93.1\%} &
  \tiny{345m (est)} &
  \tiny{9.4k-392.7k} &
  \tiny{3$\scalemath{0.7}{\times} \scalemath{0.9}{10^{\scalemath{0.8}{-5}}}$-$\scalemath{0.9}{10^{\scalemath{0.8}{-3}}}$ (est)} &
  \tiny{66.8-91.4\% acc (87.78-98.17\% rel acc); 72.5-92.8\% fid} \\ \hline
SMA-\tiny{RNN} \tiny{\cite{he_model_2021}} &
  \tiny{BERT} &
  \tiny{Same as target} &
  \tiny{Text} &
  \tiny{97.1\%} &
  \tiny{110m (est)} &
  \tiny{\unknown} &
  \tiny{\unknown} &
  \tiny{92.8\% acc 76.5\% tr} \\ \hline
\multirow{2}{0.99\linewidth}{SMA-\tiny{RNN} \tiny{\cite{takemura_model_2020}}} &
  \multirow{2}{0.99\linewidth}{\tiny{LSTM}} &
  \tiny{RNN} &
  \tiny{Image} &
  \tiny{97.3\%} &
  \multirow{2}{0.99\linewidth}{\tiny{\unknown}} &
  \multirow{2}{0.99\linewidth}{\tiny{\unknown}} &
  \multirow{2}{0.99\linewidth}{\tiny{\unknown}} &
  \tiny{90-97.5\% acc} \\ \cline{3-5} \cline{9-9} 
 &
   &
  \tiny{LSTM} &
  \tiny{Sequential} &
  \tiny{0.899$\text{\tiny{R}}^\text{\tiny{2}} $} &
   &
   &
   &
  \tiny{0.85$\text{\tiny{R}}^\text{\tiny{2}}$ (est)} \\ \hline
SMA-\tiny{GNN} \tiny{\cite{defazio_adversarial_2020}} &
  \tiny{GCN} &
  \tiny{Same as target} &
  \tiny{Graph} &
  \tiny{\unknown} &
  \tiny{\unknown} &
  \tiny{70} &
  \tiny{\unknown} &
  \tiny{64-80\% fid} \\ \hline
SMA-\tiny{GNN} \tiny{\cite{wu_model_2022}} &
  \tiny{GCN} &
  \tiny{Same as target} &
  \tiny{Graph} &
  \tiny{71.3-81.6\%} &
  \tiny{\unknown} &
  \tiny{60-120} &
  \tiny{\unknown} &
  \tiny{70.8-79.9\% acc; 84.6-89.6\% fid} \\ \hline
SMA-\tiny{GNN (ShD) \tablefootnote{Shadow data}} \tiny{\cite{wu_model_2022}} &
  \tiny{GCN} &
  \tiny{Same as target} &
  \tiny{Graph} &
  \tiny{69.7-81.6\%} &
  \tiny{\unknown} &
  \tiny{60-120} &
  \tiny{\unknown} &
  \tiny{70.8-83.2\% acc; 73.6-83.7\% fid} \\ \hline
SMA-\tiny{GNN} \tiny{\cite{he_stealing_2021}} &
  \tiny{GCN} &
  \tiny{Same as target} &
  \tiny{Graph} &
  \tiny{\unknown} &
  \tiny{\unknown} &
  \tiny{10\% of nodes} &
  \tiny{\unknown} &
  \tiny{0.958-0.999 AUC} \\ \hline
SMA-\tiny{GNN} \tiny{\cite{shen_model_2022}} &
  \tiny{GIN} &
  \tiny{GIN} &
  \tiny{Graph} &
  \tiny{92.4 \%} &
  \tiny{\unknown} &
  \tiny{5.9k (est)} &
  \tiny{\unknown} &
  \tiny{87.7\% acc, 90.6\% fid} \\ \hline
\end{tabular}%
}
\end{table}
(2) Non-available data, denoted as \unknown. It corresponds to data that is not provided by the authors of papers.
In some cases, we estimated certain data. For example, we could estimate the number of parameters of the target model from a given model architecture name. These cases, together with corresponding efficiency scores, are indicated in the table with \textit{est}. We also used this notation for attack performance scores extracted from plots or diagrams when no exact numbers were provided. 

One observation from \Cref{tab:attacks} is that a comprehensive comparison between the proposed methods is difficult due to the lack of a uniform reporting standard. We highlight two issues:

\begin{enumerate}

\item 
Effectiveness is reported in multiple ways, e.g. absolute accuracy on the original classification task (which is not comparable among different datasets).
Relative accuracy and fidelity are more expressive, as they contrast the effectiveness of the stolen model with the original one. Which of these two measures is more important depends on the exact use case. Reporting both would therefore be the preferred approach.
For a more extensive analysis, we also suggest to report the transferability rate of the stolen model since it reveals the similarity between the decision boundaries of the target- and adversary models.

\item 
It is difficult to properly assess the efficiency of several of the attacks since the literature very often omits important aspects. For example, the absolute number of queries needs to be rather put in relation to the amount of information that needs to be stolen, such as the number of learned parameters of the target model, to compute an average amount of queries required per parameter. However, in many cases at least one of the required numbers for computing the score is not provided.
\end{enumerate}

We note that the efficiency scores are most useful when comparing different attack variations on similar model types since attacks are not directly comparable across model types. As such, it might be feasible for an attacker to spend 10 queries per parameter when stealing a model with 1,000 parameters in total, but such ratios would be prohibitive if stealing a CNN model with millions of parameters.
We also want to point out that comparing attacks should take the adversary's capabilities into account. In \Cref{tab:attacks}, we highlight some settings like the difference between the target- and the adversary model architecture; however, due to limited space, we are omitting for instance the difference between original- and attacker training data.

\section{Side-channel Attacks}
\label{sec:side-channel}

\begin{table}[t]
\centering
\caption{Taxonomy and attack success of side-channels attacks. \fully indicates that the stealing goal is fully achieved, while \partially indicates partial success.}
\label{tab:side-channel}
\begin{tabular}{|l|l|l|p{4cm}|l|c|c|} 

\cline{6-7}
\multicolumn{5}{l|}{}                                                                                                                                                                                                                                                                                            & \multicolumn{2}{c|}{\small{\textbf{Success}}}                                      \\ 
\hline
\small{\textbf{Access}}            & \small{\textbf{Channel}}                                                                   & \small{\textbf{Stealing goal}}                    & \small{\textbf{Target Model}}                                                                                       & \small{\textbf{Ref}}                         & \multicolumn{1}{l|}{\small{\textbf{Arch}}} & \multicolumn{1}{l|}{\small{\textbf{Params}}}  \\ 
\hline
\multirow{5}{*}{\small{Software}}  & \small{Timing}                                                                             & \multirow{5}{*}{\small{Architecture}}             & \tiny{CNN}                                                                                                  & \small{\cite{duddu_stealing_2019}}           & \small{\partially}                         &                                       \\ 
\cline{2-2}\cline{4-7}
                           & \multirow{4}{*}{\small{Cache}}                                                             &                                           & \tiny{VGG-19, ResNet-50}                                                                                    & \small{\cite{hong_security_2020}}            & \small{\fully}                             &                                       \\ 
\cline{4-7}
                           &                                                                                    &                                           & \tiny{VGG-16, ResNet-50}                                                                                    & \small{\cite{yan_cache_2020}}                & \small{\partially}                         &                                       \\ 
\cline{4-7}
                           &                                                                                    &                                           & \tiny{AlexNet, VGG-13/16}                                                                                   & \small{\cite{liu_ganred_2020}}               & \small{\fully}                             &                                       \\ 
\cline{4-7}
                           &                                                                                    &                                           & \tiny{MalConv, ProxalessNAS}                                                                                & \small{\cite{hong_how_2020}}                 & \small{\fully}                             &                                       \\ 
\hline
\multirow{15}{*}{\small{Hardware}} & \small{PCIe Bus}                                                                           & \small{Parameters}                                & \tiny{ResNet-34/101/152,NasNet}                                                                             & \small{\cite{hu_deepsniffer_2020}}           & \small{\fully}                             &                                       \\ 
\cline{2-7}
                           & \multirow{3}{*}{\small{Memory}}                                                            & \small{Parameters}                                & \tiny{quantized CNN (ResNet-18/34, VGG-11)}                                                                 & \small{\cite{rakin_deepsteal_2022}}          &                                    & \small{\partially}                            \\ 
\cline{3-7}
                           &                                                                                    & \small{Architecture}                              & \tiny{Alexnet, VGG-16, Resnet-18/50/101}                                                                    & \small{\cite{wang_demystifying_2022}}        & \small{\partially}                         & \multicolumn{1}{l|}{}                 \\ 
\cline{3-7}
                           &                                                                                    & \small{Architecture, parameters}                  & \tiny{AlexNet, SqueezeNet}                                                                                  & \small{\cite{hua_reverse_2018}}              & \small{\fully}                             & \small{\partially}                            \\ 
\cline{2-7}
                           & \multirow{7}{*}{\begin{tabular}[c]{@{}l@{}}\small{EM} \\\small{(electro-}\\\small{magnetic)}\end{tabular}} & \multirow{2}{*}{\small{Parameters}}               & \tiny{NN}                                                                                                   & \small{\cite{yoshida_model-extraction_2019}} &                                    & \small{\partially}                            \\ 
\cline{4-7}
                           &                                                                                    &                                           & \tiny{AlexNet, VGG-16, (Wide)ResNet-50, Inception-v1/v3, DenseNet, NasNet, Xception, Inception-ResNet-50-2} & \small{\cite{breier_sniff_2021}}             &                                    & \small{\fully}                                \\ 
\cline{3-7}
                           &                                                                                    & \small{Architecture}                              & \tiny{AlexNet, VGG-19}                                                                                      & \small{\cite{hu_deepsniffer_2020}}           & \small{\fully}                             &                                       \\ 
\cline{3-7}
                           &                                                                                    & \multirow{2}{*}{\small{Architecture, parameters}} & \tiny{Binary NN}                                                                                            & \small{\cite{regazzoni_machine_2020}}        & \small{\partially}                         & \small{\partially}                            \\ 
\cline{4-7}
                           &                                                                                    &                                           & \tiny{NN, CNN}                                                                                              & \small{\cite{batina_csi_2019}}               & \small{\fully}                             & \small{\partially}                            \\ 
\cline{3-7}
                           &                                                                                    & \multirow{3}{*}{\small{Architecture, parameters}} & \tiny{Decision Tree}                                                                                        & \small{\cite{jap_practical_2020}}            & \small{\fully}                             & \small{\fully}                                \\ 
\cline{4-7}
                           &                                                                                    &                                           & \tiny{Binary NN, CNN, VGG, LeNet, AlexNet}                                                                  & \small{\cite{yu_deepem_2020}}                & \small{\partially}                         & \small{\partially}                            \\ 
\cline{2-2}\cline{4-7}
                           & \multirow{3}{*}{\small{Power trace}}                                                       &                                           & \tiny{AlexNet, Inception-v3, ResNet-50/101}                                                                 & \small{\cite{xiang_open_2020}}               & \small{\partially}                         & \small{\partially}                            \\ 
\cline{3-7}
                           &                                                                                    & \multirow{2}{*}{\small{Parameters}}               & \tiny{NN}                                                                                                   & \small{\cite{li_model_2021}}                 &                                    & \small{\partially}                            \\ 
\cline{4-7}
                           &                                                                                    &                                           & \tiny{Binarized NN}                                                                                         & \small{\cite{dubey_high-fidelity_2022}}      &                                    & \small{\partially}                            \\ 
\cline{2-7}
                           & \small{PCIe}                                                                               & \small{Architecture, parameters}                  & \tiny{NN, VGG-16, ResNet-20}                                                                                & \small{\cite{zhu_hermes_2021}}               & \small{\fully}                             & \small{\fully}                                \\
\hline
\end{tabular}

\end{table}

Side-channel (SC) attacks (SCAs) exploit hardware- or software characteristics to reveal the model. Therefore, their performance strongly depends on the device on which the target model is running.
SCAs were initially proposed for key recovery attacks in cryptography, e.g. against RSA \cite{kocher_timing_1996}. Recent usages of SCAs extend to the model stealing domain, where they have most commonly been employed to extract the model architecture; however, some attacks also target other hyperparameters or learned model parameters. 
While some authors are calling their attack a "reverse-engineering attack", we will use the terminology defined in \Cref{sec:terminology} and call it a model stealing attack. 

\Cref{tab:side-channel} provides a classification of side-channel attacks based on access to the model and the exploited channels. 
Generally speaking, for an SCA, an attacker models possible effects of specific causes, e.g. by generating a set of candidate models and observing how their inference influences the respective side-channel. If the attacker is observing effects of an unknown model on the side-channel, this can then be used to learn information about possible causes, e.g. possible model hyperparameters.

\subsection{Software Access}
While having software access to the device, an adversary can exploit cache- or timing side channels. Both channels can be used to infer the type of computational operations performed; however, attacks based on this can only extract the architecture of a model.
Most cache side channels try to manipulate the contents of a cache (shared or otherwise accessible, e.g. via cache conflicts) prior to the running of the target process to force a reload of the cache from memory. The timing difference (memory needed to be reloaded or not) can then be used to infer if the target accessed that cache. Well-known attacks of this kind are e.g. \textit{PRIME + PROBE} \cite{tromer_efficient_2010}, and \textit{FLUSH + RELOAD} \cite{yarom_flushreload_2014}

Duddu et al. \cite{duddu_stealing_2019} exploited timing side channels to extract the depth of the network. Based on this information, they evaluated a set of candidate architectures and selected the one with the most similar prediction behaviour to the target model. This attack requires a membership inference attack beforehand, as original training data is needed to evaluate (and select from) the candidate architectures. The authors applied reinforcement learning to construct the optimal substitute architecture.
Hunt et al. \cite{hunt_telekine_2020} used GPU kernel execution time for predicting classification outputs. 

Hong et al. \cite{hong_security_2020} used a Flush+Reload side channel to match an observed architecture to a set of candidate architectures by learning a Decision Tree meta-model on the SC attributes, thus demonstrating the ability to steal VGG-19 and ResNet-50 architectures out of 13 candidates.
Hong et al. \cite{hong_how_2020} also exploited a Flush+Reload cache SC to reveal a CNN architecture. They extracted the trace of calls of specific Pytorch- or Tensorflow functions that compose an NN. Observed execution times are then mapped onto a computational graph that corresponds to the target model. They demonstrated the attack performance against MalConv- \cite{raff_malware_2018} and ProxylessNAS \cite{cai_proxylessnas_2019} models.
Yan et al. used Prime+Probe and Flush+Reload to extract VGG and ResNet architectures \cite{yan_cache_2020}. The authors analysed Generalised Matrix Multiply executions and revealed, with the help of a meta-model, DNN hyperparameters responsible for the network architecture such as the kernel size and number of layers. The proposed attack allows to reduce the search space of architecture candidates. 
Liu and Srivastava \cite{liu_ganred_2020} also utilised the information leaked via cache side-channel. They introduced a framework in which DNNs are characterised by the patterns of their access to specific caches over time. Their architecture stealing attack does not require sharing the memory segment between the attacker and the model, unlike e.g. \cite{hong_security_2020}, and allows the exact architecture reconstruction (and not only restricting the search space as in \cite{yan_cache_2020}).

\subsection{Hardware Access}
Hardware access to the device on which the model is executed opens a door for more advanced attacks. These side-channel attacks are based on the observation that all computation running on a certain platform results in unintentional physical leakages. These manifest as physical signatures of reaction time, power consumption, or electromagnetic (EM) emanations while the data is manipulated.

Hua et al. \cite{hua_reverse_2018} were among the first to implement an attack using hardware-access side-channels. 
They showed that the architecture and parameters of a CNN can be revealed through the inputs and outputs of the accelerator- and off-accelerator memory access patterns, even if the accelerator has a protected memory access (in an enclave). 
Rakin et al. adapted a rowhammer memory SCA to steal parameters of a CNN quantised to 8-bit \cite{rakin_deepsteal_2022}. 
Wang et al. studied architecture extraction attacks using hardware side channels \cite{wang_demystifying_2022}. They explored how model execution events can be observed through hardware behaviour (calling these observations "Arch-hints"), which side channels can be used, and how one can estimate the effectiveness of a given Arch-hint. Then they applied their observations to launch an attack against Unified Memory, i.e. the memory is shared among all processes running on the machine to track the model traffic and use that information to extract the sequence of layers in the target model. 

Hu et al. \cite{hu_deepsniffer_2020} proposed an architecture stealing attack leveraging electromagnetic (EM) emanations or PCI-express bus events as side channel to infer read/write volume, memory addresses and execution time as features of a CNN. They learned the relation between these features and model internal architecture aspects such as CNN layer types and sizes of layers and kernels.
Batina et al. \cite{batina_csi_2019} proposed an attack for stealing architecture and parameters of NNs, extracting the activation functions, the number of layers and neurons in the layers, the number of output classes, and parameters via an EM channel.
Subsequently, their methodology was used by Jap et al. to attack tree-based algorithms \cite{jap_practical_2020}.

Yoshida et al. showed a parameter stealing attack on a DNN accelerator implemented on an FPGA (field-programmable gate array) \cite{yoshida_model-extraction_2019}. This work shows that an adversary can extract model parameters by exploiting EM leakage even if they are protected by data encryption. 
Dubey et al. considered a parameter extraction attack against a Binarised Neural Network (BNN) running on a remote multi-tenant FPGA platform via a power SC \cite{dubey_high-fidelity_2022}.  
Yu et al. \cite{yu_deepem_2020} combined side-channel and query-based approaches. They stole the architecture of a NN via an EM side-channel and then trained a substitute model using adversarial examples. 
Their research was extended by Regazzoni et al. \cite{regazzoni_machine_2020} who also studied NN structure identification via EM emanations. 
Xiang et al. \cite{xiang_open_2020} leveraged power traces to reveal the architecture of a DNN, estimated sparsity of parameters, and derived the weights. 
Zhu et al. \cite{zhu_hermes_2021} identified a new attack surface: unencrypted PCIe traffic to observe GPU-based operations. 
The proposed attack, called "Hermes Attack", succeeds in stealing a DNN model with identical hyperparameters, parameters, and architecture.
Li and Merkel investigated how the availability of power side-channel leakage can improve the transferability of a substitute model \cite{li_model_2021}. They trained a substitute model and compared its performance with a model that also uses power consumption information. The results showed that power information helps to increase the similarity between weights, but not the transferability.

Breier et al. \cite{breier_sniff_2021} introduced a parameter extraction attack on DNNs obtained via transfer learning \cite{pan_survey_2010}, i.e. a known architecture and only a few fine-tuned layers.
This is achieved through a fault injection attack, specifically an attack flipping a sign bit. The fault injection requires power- or EM leakage to detect the right time for the attack. Then, from the differences of the original output and the sign-flipped output, model parameters are reconstructed.

\section{Defences against Model Stealing} 
\label{sec:defences}

In this section, we provide an overview and systematisation of defences against model stealing attacks. \Cref{tab:defences} shows our proposed taxonomy and classifies defence approaches.

\begin{table}[ht]
\centering
\caption{Taxonomy of defence techniques.}
\label{tab:defences}

\begin{tabular}{|c|c|c|}
\hline
\small{\textbf{Defence goal}}       & \small{\textbf{Method}}           & \small{\textbf{Papers}} \\ \hline
\multirow{3}{*}{\small{Detection}}  & \small{Unique model identifier}   & \small{\cite{maini_dataset_2021, lukas_deep_2021}}         \\ \cline{2-3}
                            & \small{Watermarking}              &   \small{\cite{jia_entangled_2021, szyller_dawn_2021, chakraborty_dynamarks_2022, li_defending_2022}}                \\ \cline{2-3} 
                            & \small{Monitor-based}             & \small{\cite{kesarwani_model_2018, juuti_prada_2019, yan_monitoring-based_2021, yu_cloudleak_2020, zhang_seat_2021, pal_stateful_2021, liu_seinspect_2022, sadeghzadeh_hoda_2022, dziedzic_increasing_2022}}              \\
                            \hline
\multirow{6}{*}{\small{Prevention}} & \small{Basic}                   & \small{\cite{tramer_stealing_2016,shi_active_2018}}                \\ \cline{2-3} 
                            & \small{Re-training from scratch} & \small{\cite{atli_extraction_2020, krishna_thieves_2020}}                \\ \cline{2-3} 
                            & \small{Differential privacy}      & \small{\cite{zheng_bdpl_2019, yan_monitoring-based_2021}}                \\ \cline{2-3} 
                            & \small{Input perturbations}      & \small{\cite{grana_perturbing_2020, guiga_neural_2020, wang_information_2021, hu_stealing_2021}}                 \\ \cline{2-3} 
                            & \small{Output perturbations}     &    \small{\cite{orekondy_prediction_2020,lee_defending_2019, kariyappa_defending_2020, wang_information_2021, chen_das-ast_2020, kariyappa_protecting_2021, lee_model_2022, mazeika_how_2022, hu_stealing_2021, alabdulmohsin_adding_2014}}             \\ \cline{2-3} 
                            & \small{Model modification}        &     \small{\cite{xu_deepobfuscation_2018, lin_bident_2020, chabanne_protection_2020, szentannai_preventing_2020}}            \\ \hline
\end{tabular}

\end{table}

An important distinction between defences concerns their mode: reactive, e.g. \textit{detection} of an (ongoing or past) attack, or pro-active, i.e. \textit{prevention} of an attack.
We can further distinguish reactive defences along two goals as follows: (i) ownership verification tries to prove ownership of a stolen model and is mostly achieved by \textit{unique model identifiers} or \textit{watermarking}; it mainly aims at proving past attacks; (ii) attack \textit{detection} tries to establish whether a model is (currently) being attacked and is mostly achieved by monitoring (in itself another reactive method). 
If pro-active methods are employed, they aim to mitigate an expected attack and usually modify some aspects of the model -- the architecture, the learned parameters, the decision boundary, or the overall effectiveness of the model. An important distinction is on whether the model owner has the possibility to influence the model already during its training stage. If so, and depending on the defence asset, knowledge distillation is one strategy for instance. If the trained model is given, then various approaches can modify the output, weights, or even the architecture in a post-hoc fashion.
Reactive methods cannot prevent that a model gets stolen, but inform the model owner about the incident. However, detecting an ongoing attack via a reactive monitor might be a trigger for a pro-active defence to mitigate or even halt the attack. This might be more beneficial than applying pro-active methods upfront, since these generally also have a negative impact on legitimate users, e.g. reduced predictive accuracy.
It should be noted that a defence's success is not a binary state, i.e. completely preventing that information gets stolen or failing to do so. In many settings, it is sufficient if the defence can, for example, lower the fidelity of the stolen model so that it becomes useless for the adversary.

\subsection{Attack Detection (reactive)}

\subsubsection{Unique Model Identifier}
\label{sec:umi}
Unique model identifier (UMI) is a reactive approach to prove the ownership of a model, similar to the concept of device fingerprinting~\cite{kohno_remote_2005}. The idea is to identify a unique model property that will transfer to a substitute model during model stealing. A model owner can then verify that a model was stolen by revealing this property, but in contrast to model watermarking~(cf. \Cref{sec:defences:wm}), the owner does not need to actively embed it, as it is model-inherent.
Maini et al. proposed a defence called Dataset Inference (DI) \cite{maini_dataset_2021} which allows checking whether a model was trained on a specific dataset. The defence is based on the idea that training samples have a larger distance to the decision boundary than other samples.
The model owner can use a subset of the original training data to measure if those samples are far from the decision boundary for the substitute model. 
If it holds, the substitute model contains the identifier of the target model and, hence, can be considered as stolen. 
However, DI is not applicable if the original dataset is publicly available since other models trained independently on this dataset will be recognised as stolen. This issue was highlighted by Li et al., who showed that DI incorrectly classifies a benign model as stolen if it is trained on data that comes from the same distribution as the original data \cite{li_defending_2022}. They proposed embedding some external knowledge into the target model as an alternative solution (see \Cref{sec:defences:wm}). 
Lukas et al. aimed to find a specific subclass of transferable adversarial examples -- termed \textit{conferrable adversarial examples} -- to obtain a unique fingerprint from substitute models \cite{lukas_deep_2021}. Conferrability means that these adversarial examples transfer to substitute models, but not to other independently trained models. Hence, conferrable examples can be used to check if a particular model is a substitute for the target model. 

\subsubsection{Watermarking}
\label{sec:defences:wm}
Model watermarking (WM) is another approach to prove ownership of a (stolen) model~\cite{lederer_identifying_2023}. In contrast to UMIs, this is usually achieved by actively embedding hidden information in the model that only the legitimate owner knows how to extract. 
One possible way is to build secret backdoors into the model: during training, a model learns to predict the predefined values for some outlier samples, i.e. the model overfits to specific outliers. Then, knowing these specific samples, one can query the model and recognise the watermark through its predictions.

To resists model stealing, watermarks need to persist during the attack and appear as well in the stolen model.
Jia et al. \cite{jia_entangled_2021} trained a model to extract common features from problem-domain samples and watermarking samples. This approach guarantees that an adversary who queries the model on the problem-domain data distribution will extract watermarks together with model behaviour. 
Szyller et al. \cite{szyller_dawn_2021} proposed a strategy called DAWN -- instead of applying watermarking during the training process, they change, for a small number of queries, the output of the model, thus using queried samples as (dynamic) watermark carriers. 
Chakraborty et al. proposed another dynamic watermarking strategy called DynaMarks \cite{chakraborty_dynamarks_2022}. In contrast to DAWN, the authors alter the probabilities that the target model returns, thus in most cases preserving more utility of the model. 
The added perturbations are randomised; hence, an attacker cannot bypass this defence by querying the same sample several times. The watermark is then extracted through comparing the distributions of probabilities per class returned by the substitute model and the original protected model. 

Li et al. modified a part of the training set by applying Style-GAN while preserving the original labels \cite{li_defending_2022}. 
Then, by observing model gradients on a modified image, they were able to say if the knowledge about those samples is present in a given model. However, this defence requires white-box access for ownership verification.
A recent survey of further watermarking approaches is given by ~\cite{lederer_identifying_2023}.

\subsubsection{Monitor-based}
\label{sec:monitor-based}
Another reactive defence approach is detecting malicious users by analysing the queries. This approach is called monitor \cite{kesarwani_model_2018} or monitoring-based \cite{yan_monitoring-based_2021}. 
Kesarwani et al. \cite{kesarwani_model_2018} implemented a monitor which estimates the data space covered by the issued queries, thus inferring a kind of "extraction completeness status" (ECS). The authors used this approach to detect attacks on decision trees. 
Juuti et al. \cite{juuti_prada_2019} proposed a defence technique named PRADA which analyses the distribution of queried samples. Their method is based on detecting a deviation from the normal distribution in the distances between queried samples.

Yu et al. proposed a monitor called DefenseNet \cite{yu_cloudleak_2020}, which is an NN trained to classify if a sample is adversarial or benign. As input features, DefenseNet takes all outputs from each hidden layer of the target model produced during sample forward propagation. 
Zhang et al. introduced SEAT, a monitor that aims to defend against attacks using adversarial examples \cite{zhang_seat_2021}. They trained an encoder that checks whether a current query is too close to any of the previous queries and, thus, likely to be an adversarial example. 
As soon as the number of such detections exceeds a certain threshold, the corresponding user is blocked. 
Pal et al. proposed a monitor called VarDetect which uses a variational autoencoder (VAE) \cite{pal_stateful_2021}. The monitor collects queries, and if the count reaches a certain number, the monitor checks if those queries are coming from a benign (original or PD data) or malicious (artificial, adversarial PD or NPD data) client. 
Liu et al. proposed SeInspect, a two-stage monitor for image data that first analyses the last batch of queries sent by a user and, if it seems suspicious, analyses the user’s whole query history to detect an attack \cite{liu_seinspect_2022}. The authors showed that although a slightly perturbed image can be indistinguishable from the original, the features for these images on the penultimate layer of the target network differ significantly. 
They used this observation to detect both adversarial examples and NPD images as malicious queries.

Sadeghzadeh et al. utilised a notion of the hardness of data samples to launch a monitor called HODA \cite{sadeghzadeh_hoda_2022}. 
The hardness of a sample is determined by the number of epochs required for its prediction to stabilise.
The authors showed that in-distribution samples are generally easier to learn than NPD or adversarial examples, so the latter can be detected by measuring their hardness scores.
Dziedzic et al. designed a monitor-based defence that increases the effort to query hard examples based on ideas of the Proof-of-Work (PoW) principle \cite{dziedzic_increasing_2022}. They exploited differential privacy techniques to quantitatively measure the extracted information of a query and created a proportionally difficult puzzle the attacker needs to solve before the query is answered, effectively slowing down the attacker's querying process.

\subsection{Attack Prevention (proactive)}
Attack prevention approaches are mostly directed against the effectiveness of the attack, i.e. they do not prevent the attacker from obtaining a model, but aim to render the quality of the stolen model too low for it to be useful.

\subsubsection{Basic Defences}
This section covers basic defence approaches, all of which are based on simple ideas and can be easily implemented.
Tramèr et al. \cite{tramer_stealing_2016} proposed defences with low implementation overhead that can be applied to models that return detailed class prediction information, such as soft-max outputs or logits. One approach is to return only the label predicted for the sample, but no additional information. 
\subsubsection{Training from Scratch}
Atli et al. \cite{atli_extraction_2020} explored defence techniques against Knockoff nets \cite{orekondy_knockoff_2019} (cf. \Cref{sec:attack:query:substitute}). In the original version of the attack, the target models were pre-trained on ImageNet; ImageNet was also used by the attacker to query the target model. Atli et al. trained two models with a specific architecture for the data domain from scratch; attacking them with Knockoff nets was then less successful.
A similar approach was explored by Krishna et al. \cite{krishna_thieves_2020} as a countermeasure against their attack on BERT-based models. They trained a model from scratch \cite{pan_survey_2010} and observed that the F1 score of the stolen model had also decreased.

\subsubsection{Data perturbation} \label{sec:defences:prevention:perturbation}
Several studies showed how data perturbation can be used to defend against model stealing (see \Cref{tab:defences}). The main idea of this approach is to make the model predict an inexact output while preserving integrity. There are two types of data to perturb: data fed to the model, i.e. input, or the model prediction, i.e. output.

Wang et al. introduced the concept of Information Laundering (IL) \cite{wang_information_2021}. They considered input and output perturbations simultaneously to achieve two goals: (i) hide the predictions of the target model to increase its confidentiality, and (ii) preserve the utility of the model. The work theoretically describes the optimal distribution of input and output perturbations such that both goals are achieved. 

\paragraph{Input Perturbation (IP)}
Grana analytically proved that perturbations added to inputs prevent logistic- and linear regressions from parameter stealing \cite{grana_perturbing_2020}.
Guiga and Roscoe \cite{guiga_neural_2020} protect image models by adding noise to the unimportant pixels selected by the Gradient-weighted Class Activation Mapping (Grad-CAM) method \cite{selvaraju_grad-cam_2017}.
Hu and Pang proposed two defences for GAN protection \cite{hu_stealing_2021}. The first defence takes several input queries and replaces each of them with an input obtained as a result of linear interpolation of two original queries. The second defence applies constraints on inputs such that outputs can belong only to a predefined set (e.g., generating faces with only green eyes). 

\paragraph{Output Perturbation (OP)}
Tramèr et al. \cite{tramer_stealing_2016} provided a basic form of this defence, namely rounding the predicted scores; they state, however, that it is not a promising strategy.
A more advanced form of OP was proposed by Orekondy et al. \cite{orekondy_prediction_2020} who named their defence Maximising Angular Deviation (MAD). The main idea is to perturb output confidence scores to make the gradient maximally far from the original. 
Lee et al. \cite{lee_defending_2019} proposed to use the reverse sigmoid activation function as a defence. A specific characteristic of that function is that it maps different logit values to the same probability. This leads to wrong gradient values and complicates the stealing process.

Shi et al. \cite{shi_evasion_2017} proposed a defence against decision boundary stealing.
They claim that flipping some labels in the training set can make the model more robust against evasion attacks and, thus, against revealing the decision boundary. 
Kariyappa and Qureshi \cite{kariyappa_defending_2020} proposed a more advanced approach than Shi et al. \cite{shi_evasion_2017}: wrong predictions are returned only for queries that are out of distribution.
Chen et al. utilised an adaptive softmax transformation to perform another OP defence called DAS-AST \cite{chen_das-ast_2020}. By modifying softmax outputs, they changed the distribution of samples obtained by an attacker, misleading them from the decision boundary. 

One of the earliest defences was reported by Alabdulmohsin et al. in 2014 \cite{alabdulmohsin_adding_2014}, when the authors explored the security of SVMs against several types of adversarial attacks. They proposed to train several models and randomly pick one of them to produce an output. 
Kariyappa et al. extend on that by a defence called Ensemble of Diverse Models (EDM) \cite{kariyappa_protecting_2021}. They train multiple models that all accurately classify in-distribution samples, but are trained to on purpose predict diverse values for out-of-distribution samples. 
Since the decision boundaries of those models different, the accuracy of a substitute model trained on out-of-distribution data is decreased. 

Lee et al. considered an OP defence called DeepDefence for target models that return probabilities and gradient-based explanations in the form of attribution maps (such as Grad-CAM) \cite{lee_model_2022}. They proposed to perturb the gradients of the target model to make them orthogonal to the original ones, while preserving the order of the top-k probabilities and the values of an attribution map. 
Mazeika et al. argued that some highly-effective OP defences perturb confidence scores too much, thus harming benign users \cite{mazeika_how_2022}. In contrast, they introduced a defence called GRAD$^2$, which adds perturbations that direct the substitute model training in a predefined (non-optimal) direction while preserving the number of total changes below a certain threshold. 

Hu and Pang devised a concept of OP for GANs \cite{hu_stealing_2021}. They proposed to add noise to images, apply the Gaussian filter, and JPEG compression. 

\paragraph{Differential Privacy} 
\label{sec:defences:prevention:perturbation:dp}

Zheng et al. \cite{zheng_bdpl_2019} used a form of differential privacy \cite{dwork_differential_2006} against behaviour stealing.
Their main idea is to make outputs of all samples that are close to the decision boundary indistinguishable from each other.
This is achieved by adding perturbations to these outputs through a so-called "boundary differential privacy layer" (BDPL).
Yan et al. \cite{yan_monitoring-based_2021} broke BDPL with their query-flooding parameter duplication attack (QPD) (cf. \Cref{sec:ESA}) and, subsequently, proposed a new defence called MDP which combines differential privacy (as in \cite{zheng_bdpl_2019}) with monitoring to mitigate the QPD attack. 
If an attack is assumed by a monitor, the amount of noise that should be added to the data is dynamically determined.
This makes the perturbation less predictable and, thus, determining the true output more difficult.

\subsubsection{Model Modification}
Contrary to perturbing the data, which aims at reducing the precision of the stolen models' behaviour, one can modify the model architecture and/or -parameters.
The motivation for protecting architectures can e.g. be that an architecture is novel and has certain advantages over others.
The main goal of the defender is thus not to protect one specific trained instance of this architecture (i.e. the learned model parameters) or training hyperparameters, but the general architecture itself, as this should prevent an attacker to apply it to a different domain.

Xu et al. \cite{xu_deepobfuscation_2018} proposed a defence that simulated a CNN feature extractor using a shallow sequential convolutional block and used it to train a smaller model with similar performance, applying ideas from Knowledge Distillation.
Lin et al. \cite{lin_bident_2020} proposed a strategy inspired by secret sharing in cryptography. A model is first transformed into what they refer to as a \textit{bident model structure}, i.e. a model with two independent branches which merge before the output layer. Each sub-model receives the same input, and the sub-models' outputs are merged into a single output. Given the output and one sub-model, it is impossible to reconstruct another sub-model and, therefore, the whole model.
Chabanne et al. \cite{chabanne_protection_2020} investigated a defence against recovering attacks (\Cref{sec:recovering}). The authors added redundant layers to a CNN with ReLU activation functions that do not change the functionality but make the model more complex and, thus, more difficult to steal. 
They further showed that the modified model's decision boundary differs from the original model's decision boundaries, but keeps a similar functionality. %
Szentannai et al. \cite{szentannai_preventing_2020} implemented a defence for NNs with fully connected layers. The authors proposed to transform a model into a functionally equivalent model, but with so-called "sensitive weights" which make the model less robust and, thus, behaviour stealing more difficult. To do this, they added deceptive neurons to the network which add noise to individual layers, but cancel each other out in the overall effect.

\section{Analysis of Attacks and Defences}\label{sec:defences-vs-attacks}
In this section, we analyse how attacks and defences compare against each other. To this end, we propose two guidelines: (1) how to steal a model, and (2) how to protect it against model stealing. Further, based on results reported in the literature, we show how model stealing attacks and defences fare against each other.

In related work, Duddu and Vijay Rao \cite{duddu_quantifying_2020} proposed a framework based on a Bayesian network to quantitatively estimate information extracted from a target DNN model through model stealing attacks. 
Based on the results, the most prominent combinations are equation-solving attacks with either a meta-model attack or a side-channel attack. It is worth mentioning that this analysis did not explore if these combinations are possible in practice. For instance, the meta-model attack  was applied for CNNs, whereas none of the equation-solving attacks targets CNNs. 

\subsection{Guideline on How to Steal a Model}
As a summary of our attacks analysis, in \Cref{fig:attacks} we provide a guideline for the best attack based on the stealing objectives and attacker's capabilities.
Depending on the objective of the attacker, an exact (for architecture, training hyperparameter, or learned parameter stealing) or approximate (for the level of effectiveness or prediction consistency stealing) attack approach is to be chosen.
\begin{figure*}[t]
\centering
\resizebox{0.9\textwidth}{!}{\input{fig-AttacksDiagram}}
\caption{A comprehensive taxonomy of model stealing attacks, in the form of an "attacker's guide". Abbreviations used: \\
SMA - Substitute Model Attack, SCA - Side-Channel Attack, MMA - Meta-Model Attack, RA - Recovering Attack, PFA - Path-Finding Attack, WFA - Witness-Finding Attack, ESA-(H)P - Equation-Solving Attack - (Hyper)Parameters} 
\label{fig:attacks}
\end{figure*}
Approximate extraction does not necessarily require information on the target model type, given that model-agnostic generic approaches, denoted as \textit{SMA*} are available. Having more detailed information on the domain or type of data can give an indication on which specific model type is well suited. Then, a method specifically designed to steal e.g. recurrent neural networks (\textit{SMA-RNN}) can be employed.
In the group of exact extraction attacks, stealing training hyperparameters requires knowledge of the model type and learned parameters. With this information available, attacks addressing specific model types such as logistic regression or SVM can be carried out.
Similarly, for stealing model parameters, attacks specialised in specific model types have been proposed, e.g. witness-finding attacks (\textit{WFA}) target linear binary model types, and path-finding attacks can be used to steal Decision or Regression Trees.
Stealing the architecture mostly applies to neural networks, where hyperparameters define the layers, neurons, activation functions, etc. 
It can be achieved in two ways:
If the attacker can issue (black-box) queries to the model, a meta-model attack, which builds on a knowledge base of known architectures, can be used. 
Such a knowledge base needs to cover many different architectures, and is thus expensive to obtain.
If such an approach is not feasible, but a hardware- or software side-channel access is available, this can be utilised.

\subsection{Guideline on How to Protect a Model}
\begin{figure*}[ht]
\centering
\resizebox{0.9\textwidth}{!}{\input{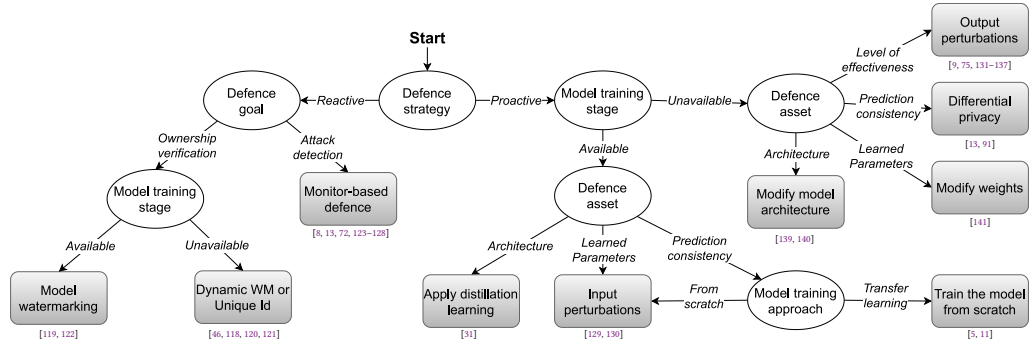}}

\caption{Comprehensive taxonomy of model stealing defences, in the form of a "model protection guideline"}
\label{fig:defences}
\end{figure*}

\Cref{fig:defences} provides a guideline on choosing a defence strategy according to particular goals and conditions. Hence, the considered defence taxonomy (\Cref{tab:defences}) has two branches on the top level: reactive and pro-active defences.
Depending on the availability of the model training stage, there are different approaches how a model owner can mitigate an attack that targets a certain asset.
For instance, if the owner wants to defend the architecture of the model, defences that modify the target model architecture are the best option since they "hide" the original architecture.
If the owner's primary goal is to track or detect malicious users of an API, then unique model identifier, model watermarking or monitor defences can be applied. However, if an adversary never makes a stolen model public, unique model identifiers and model watermarking are useless. Further, monitors may detect an ongoing attack too slowly and issue a warning of potential danger only when the target model has already been stolen. 
A combination of different defence techniques could lead to a better protection level. Following the guideline, one can choose suitable defences and combine them into a potentially more powerful defence. 

\subsection{Attacks and Defences Lineup}
\begin{table*}[t]
\centering
\caption{Comparison of attacks and reactive defences. $\attackMitigatedMark$ - attack is mitigated; \attackLoweredMark - defence has limited effect; $\defenceBrokenMark$ - defence is broken; \attackClaimMark - claim made by the authors; \speculatemark - speculation made by the authors}
\label{tab:comparison-reactive}
\resizebox{\textwidth}{!}{{
\setlength{\tabcolsep}{1pt}
\begin{tabular}{|l|c|c|c|c|c|c|c|c|c|c|c|c|}
\hline
\multicolumn{1}{|c|}{\multirow{2}{*}{\small{\textbf{Attack/Defence types}}}}                 & \multicolumn{8}{c|}{\small{\textbf{Monitor}}}                                                                                                                                                                                                                                                                                                                                                                                                                                                                                                                                                                                                                                       & \multicolumn{2}{c|}{\small{\textbf{WM}}}                                                                                                           & \multicolumn{2}{c|}{\small{\textbf{UMI}}}                                                                                                                    \\ \cline{2-13} 
\multicolumn{1}{|c|}{}                                                               & \begin{tabular}[c]{@{}c@{}}\small{ECS} \tiny{\cite{kesarwani_model_2018}}\\\end{tabular} & \begin{tabular}[c]{@{}c@{}}\small{PRADA} \tiny{\cite{juuti_prada_2019}}\\\end{tabular}                               & \begin{tabular}[c]{@{}c@{}}\small{DefenseNet} \tiny{\cite{yu_cloudleak_2020}}\\\end{tabular}    & \begin{tabular}[c]{@{}c@{}}\small{SEAT} \tiny{\cite{zhang_seat_2021}}\\\end{tabular}          & \begin{tabular}[c]{@{}c@{}}\small{VarDetect} \tiny{\cite{pal_stateful_2021}}\\\end{tabular}     & \begin{tabular}[c]{@{}c@{}}\small{PoW} \tiny{\cite{dziedzic_increasing_2022}}\\\end{tabular}                          & \begin{tabular}[c]{@{}c@{}}\small{SeInspect} \tiny{\cite{liu_seinspect_2022}}\\\end{tabular}     & \begin{tabular}[c]{@{}c@{}}\small{HODA} \tiny{\cite{sadeghzadeh_hoda_2022}}\\\end{tabular}          & \begin{tabular}[c]{@{}c@{}}\small{DAWN} \tiny{\cite{szyller_dawn_2021}}\\\end{tabular}             & \tiny{\cite{jia_entangled_2021}}                            & \multicolumn{1}{c|}{\tiny{\cite{lukas_deep_2021}}}                      & \begin{tabular}[c]{@{}c@{}}\small{DI} \tiny{\cite{maini_dataset_2021}}\\\end{tabular}                                \\ \hline
\small{ESA-P} \tiny{(dupl. quer.)} \tiny{\cite{yan_monitoring-based_2021}}                   & \multicolumn{1}{c|}{\small{\attackLoweredMark} \tiny {\cite{yan_monitoring-based_2021}}}              & \multicolumn{1}{c|}{}                                                                           & \multicolumn{1}{c|}{}                                                     & \multicolumn{1}{c|}{}                                                   & \multicolumn{1}{c|}{}                                                     & \multicolumn{1}{c|}{}                                                                           & \multicolumn{1}{c|}{}                                                      &                                                          & \multicolumn{1}{c|}{}                                                        &                                                             & \multicolumn{1}{c|}{}                                                   &                                                                            \\ \hline
\small{PFA} \tiny{\cite{tramer_stealing_2016}}                                               & \multicolumn{1}{c|}{\small{\attackMitigatedMark} \tiny{\cite{kesarwani_model_2018}}}                  & \multicolumn{1}{c|}{}                                                                           & \multicolumn{1}{c|}{}                                                     & \multicolumn{1}{c|}{}                                                   & \multicolumn{1}{c|}{}                                                     & \multicolumn{1}{c|}{}                                                                           & \multicolumn{1}{c|}{}                                                      &                                                          & \multicolumn{1}{c|}{}                                                        &                                                             & \multicolumn{1}{c|}{}                                                   &                                                                            \\ \hline
\small{SMA}-* \tiny{(retraining)} \tiny{\cite{tramer_stealing_2016, reith_efficiently_2019}} & \multicolumn{1}{c|}{}                                                                         & \multicolumn{1}{c|}{\small{\attackMitigatedMark} \tiny{\cite{juuti_prada_2019}}}                        & \multicolumn{1}{c|}{}                                                     & \multicolumn{1}{c|}{}                                                   & \multicolumn{1}{c|}{\small{\attackMitigatedMark} \tiny{\cite{pal_stateful_2021}}} & \multicolumn{1}{c|}{}                                                                           & \multicolumn{1}{c|}{\small{\attackMitigatedMark} \tiny{\cite{liu_seinspect_2022}}} &                                                          & \multicolumn{1}{c|}{}                                                        &                                                             & \multicolumn{1}{c|}{}                                                   &                                                                            \\ \hline
\small{SMA-NN} \tiny{\cite{papernot_practical_2017}}                                         & \multicolumn{1}{c|}{}                                                                         & \multicolumn{1}{c|}{\small{\attackMitigatedMark} \tiny{\cite{juuti_prada_2019}}}                        & \multicolumn{1}{c|}{}                                                     & \multicolumn{1}{c|}{\small{\attackMitigatedMark} \tiny{\cite{zhang_seat_2021}}} & \multicolumn{1}{c|}{\small{\attackMitigatedMark} \tiny{\cite{pal_stateful_2021}}} & \multicolumn{1}{c|}{\small{\attackMitigatedMark} \tiny{\cite{dziedzic_increasing_2022}}}                & \multicolumn{1}{c|}{\small{\attackMitigatedMark} \tiny{\cite{liu_seinspect_2022}}} & \small{\attackMitigatedMark} \tiny{\cite{sadeghzadeh_hoda_2022}} & \multicolumn{1}{c|}{}                                                        &                                                             & \multicolumn{1}{c|}{\small{\attackMitigatedMark} \tiny{\cite{lukas_deep_2021}}} &                                                                            \\ \hline
\small{SMA-CNN} \tiny{Copycat} \tiny{\cite{correia-silva_copycat_2018}}                      & \multicolumn{1}{c|}{}                                                                         & \multicolumn{1}{c|}{}                                                                           & \multicolumn{1}{c|}{}                                                     & \multicolumn{1}{c|}{}                                                   & \multicolumn{1}{c|}{\small{\attackMitigatedMark} \tiny{\cite{pal_stateful_2021}}} & \multicolumn{1}{c|}{\small{\attackMitigatedMark} \tiny{\cite{dziedzic_increasing_2022}}}                & \multicolumn{1}{c|}{}                                                      &                                                          & \multicolumn{1}{c|}{}                                                        &                                                             & \multicolumn{1}{c|}{}                                                   &                                                                            \\ \hline
\small{SMA-CNN} \tiny{(PD, AL)}  \tiny{\cite{pal_framework_2019}}                            & \multicolumn{1}{c|}{\small{\attackMitigatedMark} {\tiny \cite{pal_framework_2019}} \small{\speculatemark}}     & \multicolumn{1}{c|}{\small{\defenceBrokenMark} {\tiny \cite{pal_framework_2019}} \small{\speculatemark}}         & \multicolumn{1}{c|}{}                                                     & \multicolumn{1}{c|}{}                                                   & \multicolumn{1}{c|}{}                                                     & \multicolumn{1}{c|}{}                                                                           & \multicolumn{1}{c|}{}                                                      &                                                          & \multicolumn{1}{c|}{}                                                        &                                                             & \multicolumn{1}{c|}{}                                                   &                                                                            \\ \hline
\small{SMA-CNN} \tiny{(DS)} \tiny{\cite{yuan_es_2022}}                                       & \multicolumn{1}{c|}{}                                                                         & \multicolumn{1}{c|}{\small{\defenceBrokenMark} \tiny{\cite{yuan_es_2022}}}                              & \multicolumn{1}{c|}{}                                                     & \multicolumn{1}{c|}{}                                                   & \multicolumn{1}{c|}{}                                                     & \multicolumn{1}{c|}{}                                                                           & \multicolumn{1}{c|}{}                                                      &                                                          & \multicolumn{1}{c|}{}                                                        &                                                             & \multicolumn{1}{c|}{}                                                   &                                                                            \\ \hline
\small{SMA-CNN;RNN} \tiny{(Activethief)} \tiny{\cite{pal_activethief_2020}}                  & \multicolumn{1}{c|}{\small{\defenceBrokenMark} {\tiny \cite{pal_activethief_2020}} \small{\attackClaimMark}}   & \multicolumn{1}{c|}{\small{\defenceBrokenMark} {\tiny \cite{pal_activethief_2020}}}                     & \multicolumn{1}{c|}{}                                                     & \multicolumn{1}{c|}{}                                                   & \multicolumn{1}{c|}{\small{\attackMitigatedMark} \tiny{\cite{pal_stateful_2021}}} & \multicolumn{1}{c|}{}                                                                           & \multicolumn{1}{c|}{}                                                      &                                                          & \multicolumn{1}{c|}{}                                                        &                                                             & \multicolumn{1}{c|}{}                                                   &                                                                            \\ \hline
\small{SMA-CNN} \tiny{\cite{juuti_prada_2019}}                                               & \multicolumn{1}{c|}{}                                                                         & \multicolumn{1}{c|}{\small{\defenceBrokenMark} \tiny{\cite{juuti_prada_2019}}}                          & \multicolumn{1}{c|}{}                                                     & \multicolumn{1}{c|}{\small{\attackMitigatedMark} \tiny{\cite{zhang_seat_2021}}} & \multicolumn{1}{c|}{\small{\attackMitigatedMark} \tiny{\cite{pal_stateful_2021}}} & \multicolumn{1}{c|}{}                                                                           & \multicolumn{1}{c|}{\small{\attackMitigatedMark} \tiny{\cite{liu_seinspect_2022}}} & \small{\attackMitigatedMark} \tiny{\cite{sadeghzadeh_hoda_2022}} & \multicolumn{1}{c|}{\small{\attackMitigatedMark} \tiny{\cite{szyller_dawn_2021}}}    &                                                             & \multicolumn{1}{c|}{}                                                   &                                                                            \\ \hline
\small{SMA-CNN} \tiny{(Knockoff)} \tiny{\cite{orekondy_knockoff_2019}}                       & \multicolumn{1}{c|}{}                                                                         & \multicolumn{1}{c|}{}                                                                           & \multicolumn{1}{c|}{}                                                     & \multicolumn{1}{c|}{}                                                   & \multicolumn{1}{c|}{\small{\attackMitigatedMark} \tiny{\cite{pal_stateful_2021}}} & \multicolumn{1}{c|}{\small{\attackMitigatedMark} \tiny{\cite{dziedzic_increasing_2022}}}                & \multicolumn{1}{c|}{\small{\attackMitigatedMark} \tiny{\cite{liu_seinspect_2022}}} & \small{\attackMitigatedMark} \tiny{\cite{sadeghzadeh_hoda_2022}} & \multicolumn{1}{c|}{\small{\attackMitigatedMark} \tiny{\cite{szyller_dawn_2021}}}    &                                                             & \multicolumn{1}{c|}{\small{\attackMitigatedMark} \tiny{\cite{lukas_deep_2021}}} &                                                                            \\ \hline
\small{SMA-RNN} \tiny{\cite{krishna_thieves_2020}}                                           & \multicolumn{1}{c|}{}                                                                         & \multicolumn{1}{c|}{}                                                                           & \multicolumn{1}{c|}{}                                                     & \multicolumn{1}{c|}{}                                                   & \multicolumn{1}{c|}{}                                                     & \multicolumn{1}{c|}{}                                                                           & \multicolumn{1}{c|}{}                                                      &                                                          & \multicolumn{1}{c|}{\small{\attackMitigatedMark} \tiny{\cite{krishna_thieves_2020}}} &                                                             & \multicolumn{1}{c|}{}                                                   &                                                                            \\ \hline
\small{SMA-CNN} \tiny{(FeatureFool)} \tiny{\cite{yu_cloudleak_2020}}                         & \multicolumn{1}{c|}{}                                                                         & \multicolumn{1}{c|}{\small{\defenceBrokenMark} \tiny{\cite{yu_cloudleak_2020}}}                         & \multicolumn{1}{c|}{\small{\attackMitigatedMark} \tiny{\cite{yu_cloudleak_2020}}} & \multicolumn{1}{c|}{\small{\attackMitigatedMark} \tiny{\cite{zhang_seat_2021}}} & \multicolumn{1}{c|}{}                                                     & \multicolumn{1}{c|}{}                                                                           & \multicolumn{1}{c|}{}                                                      &                                                          & \multicolumn{1}{c|}{}                                                        &                                                             & \multicolumn{1}{c|}{}                                                   &                                                                            \\ \hline
\small{SMA-CNN} \tiny{(DFME)} \tiny{\cite{truong_data-free_2021}}                            & \multicolumn{1}{c|}{}                                                                         & \multicolumn{1}{c|}{}                                                                           & \multicolumn{1}{c|}{}                                                     & \multicolumn{1}{c|}{\small{\attackMitigatedMark} \tiny{\cite{zhang_seat_2021}}} & \multicolumn{1}{c|}{}                                                     & \multicolumn{1}{c|}{\small{\attackMitigatedMark} \tiny{\cite{dziedzic_increasing_2022}}}                & \multicolumn{1}{c|}{}                                                      &                                                          & \multicolumn{1}{c|}{}                                                        &                                                             & \multicolumn{1}{c|}{}                                                   &                                                                            \\ \hline
\small{SMA-Encoder} \tiny{\cite{dziedzic_difficulty_2022}}                                   & \multicolumn{1}{c|}{}                                                                         & \multicolumn{1}{c|}{\small{\attackLoweredMark} {\tiny \cite{dziedzic_difficulty_2022}} \small{\attackClaimMark}} & \multicolumn{1}{c|}{}                                                     & \multicolumn{1}{c|}{}                                                   & \multicolumn{1}{c|}{}                                                     & \multicolumn{1}{c|}{\small{\attackLoweredMark} {\tiny \cite{dziedzic_difficulty_2022}} \small{\attackClaimMark}} & \multicolumn{1}{c|}{}                                                      &                                                          & \multicolumn{1}{c|}{}                                                        & \small{\attackMitigatedMark} \tiny{\cite{dziedzic_difficulty_2022}} & \multicolumn{1}{c|}{}                                                   & \small{\attackLoweredMark} {\tiny \cite{dziedzic_difficulty_2022}} \small{\attackClaimMark} \\ \hline
\end{tabular}
}}

\end{table*}

\begin{table*}[t]
\centering
\caption{Comparison of attacks and pro-active defences. $\attackMitigatedMark$ - attack is mitigated; \attackLoweredMark - defence has limited effect; $\defenceBrokenMark$ - defence is broken; \attackClaimMark - claim made by the authors; \speculatemark - speculation made by the authors}
\label{tab:comparison-proactive}
\resizebox{\textwidth}{!}{{
\setlength{\tabcolsep}{1pt}
\begin{tabular}{|l|c|c|c|c|c|c|c|c|c|c|c|c|c|c|c|}
\hline
\multicolumn{1}{|c|}{\multirow{2}{*}{\small{\textbf{Attack/Defence types}}}}                                                  & \small{\textbf{Basic}}                                                                                                                    & \multicolumn{9}{c|}{\small{\textbf{Data Perturbation}}}                                                                                                                                                                                                                                                                                                                                                                                                                                                                                                                                                                                                                                                                                                                & \multicolumn{2}{c|}{\small{\textbf{Diff.} \textbf{Privacy}}}                                                                                             & \small{\textbf{No TL}}                                           & \multicolumn{2}{c|}{\small{\textbf{Model} \textbf{Modification}}}                                                                                                                \\ \cline{2-16} 
\multicolumn{1}{|c|}{}                                                                                                & \tiny{\cite{tramer_stealing_2016}}                                                                                                & \begin{tabular}[c]{@{}c@{}}\small{IP} \tiny{\cite{guiga_neural_2020}}\\\end{tabular}                    & \begin{tabular}[c]{@{}c@{}}\small{OP} \tiny{\cite{lee_defending_2019}}\\\end{tabular}                                      & \begin{tabular}[c]{@{}c@{}c@{}}\small{OP} \tiny{\cite{kariyappa_defending_2020}}\\\end{tabular}                     &\begin{tabular}[c]{@{}c@{}c@{}}\small{OP}\\\small{(MAD)} \tiny{\cite{orekondy_prediction_2020}}\\\end{tabular}                              & \begin{tabular}[c]{@{}c@{}c@{}}\small{OP}\\\small{(DAS-AST)} \tiny{\cite{chen_das-ast_2020}}\\\end{tabular}           & \begin{tabular}[c]{@{}c@{}c@{}}\small{OP}\\\small{(EDM)} \tiny{\cite{kariyappa_protecting_2021}}\\\end{tabular}               & \begin{tabular}[c]{@{}c@{}c@{}}\small{IP, OP} \\\tiny{\cite{hu_stealing_2021}}\\\end{tabular}                & \begin{tabular}[c]{@{}c@{}c@{}}\small{IP+OP}\\\small{(IL)} \tiny{\cite{wang_information_2021}}\\\end{tabular}             &\begin{tabular}[c]{@{}c@{}c@{}}\small{OP}\\\small{(GRAD$^2$)} \tiny{\cite{mazeika_how_2022}}\\\end{tabular}          & \begin{tabular}[c]{@{}c@{}}\small{BDPL} \\ \tiny{\cite{zheng_bdpl_2019}}\\\end{tabular}                         & \begin{tabular}[c]{@{}c@{}} \small{MDP} \tiny{\cite{yan_monitoring-based_2021}}\\\end{tabular}                   & \tiny{\cite{krishna_thieves_2020, atli_extraction_2020}} & \begin{tabular}[c]{@{}c@{}}\small{Distill.}\\\tiny{\cite{xu_deepobfuscation_2018}}\\\end{tabular}                      & \begin{tabular}[c]{@{}c@{}}\small{Parasatic}\\\tiny{\cite{chabanne_protection_2020}}\\\end{tabular}                              \\ \hline
\small{WFA} \tiny{\cite{lowd_adversarial_2005, tramer_stealing_2016, reith_efficiently_2019}}                                 &                                                                                                                                   & \multicolumn{1}{c|}{}                                                      & \multicolumn{1}{c|}{}                                                                         & \multicolumn{1}{c|}{}                                                                         & \multicolumn{1}{c|}{}                                                                            & \multicolumn{1}{c|}{}                                                     & \multicolumn{1}{c|}{}                                                             & \multicolumn{1}{c|}{}                                                    & \multicolumn{1}{c|}{}                                                         &                                                     & \multicolumn{1}{c|}{\small{\attackMitigatedMark} {\tiny \cite{zheng_bdpl_2019}}}         &                                                               &                                                          & \multicolumn{1}{c|}{}                                                                    &                                                                               \\ \hline
\small{ESA-P} \tiny{\cite{tramer_stealing_2016, reith_efficiently_2019}}                                                      & \small{\defenceBrokenMark} {\tiny \cite{tramer_stealing_2016}}  \small{\defenceBrokenMark} {\tiny \cite{reith_efficiently_2019}} \small{\attackClaimMark} & \multicolumn{1}{c|}{\small{\attackMitigatedMark} {\tiny \cite{guiga_neural_2020}}} & \multicolumn{1}{c|}{}                                                                         & \multicolumn{1}{c|}{}                                                                         & \multicolumn{1}{c|}{}                                                                            & \multicolumn{1}{c|}{}                                                     & \multicolumn{1}{c|}{}                                                             & \multicolumn{1}{c|}{}                                                    & \multicolumn{1}{c|}{\small{\attackMitigatedMark} \tiny{\cite{wang_information_2021}}} &                                                     & \multicolumn{1}{c|}{}                                                            & \small{\attackMitigatedMark} {\tiny \cite{yan_monitoring-based_2021}} &                                                          & \multicolumn{1}{c|}{}                                                                    &                                                                               \\ \hline
\small{ESA-P} \tiny{(dupl. quer.)} \tiny{\cite{yan_monitoring-based_2021}}                                                    & \small{\defenceBrokenMark} {\tiny \cite{yan_monitoring-based_2021}}                                                                       & \multicolumn{1}{c|}{}                                                      & \multicolumn{1}{c|}{}                                                                         & \multicolumn{1}{c|}{}                                                                         & \multicolumn{1}{c|}{}                                                                            & \multicolumn{1}{c|}{}                                                     & \multicolumn{1}{c|}{}                                                             & \multicolumn{1}{c|}{}                                                    & \multicolumn{1}{c|}{}                                                         &                                                     & \multicolumn{1}{c|}{\small{\defenceBrokenMark} {\tiny \cite{yan_monitoring-based_2021}}} & \small{\attackMitigatedMark} {\tiny \cite{yan_monitoring-based_2021}} &                                                          & \multicolumn{1}{c|}{}                                                                    &                                                                               \\ \hline
\small{PFA} \tiny{\cite{tramer_stealing_2016}}                                                                                & \small{\defenceBrokenMark} {\tiny \cite{tramer_stealing_2016}}                                                                            & \multicolumn{1}{c|}{}                                                      & \multicolumn{1}{c|}{}                                                                         & \multicolumn{1}{c|}{}                                                                         & \multicolumn{1}{c|}{}                                                                            & \multicolumn{1}{c|}{}                                                     & \multicolumn{1}{c|}{}                                                             & \multicolumn{1}{c|}{}                                                    & \multicolumn{1}{c|}{}                                                         &                                                     & \multicolumn{1}{c|}{}                                                            &                                                               &                                                          & \multicolumn{1}{c|}{}                                                                    &                                                                               \\ \hline
\small{RA}  \tiny{\cite{milli_model_2019, jagielski_high_2020, rolnick_reverse-engineering_2020, carlini_cryptanalytic_2020}} &                                                                                                                                   & \multicolumn{1}{c|}{}                                                      & \multicolumn{1}{c|}{}                                                                         & \multicolumn{1}{c|}{}                                                                         & \multicolumn{1}{c|}{}                                                                            & \multicolumn{1}{c|}{}                                                     & \multicolumn{1}{c|}{}                                                             & \multicolumn{1}{c|}{}                                                    & \multicolumn{1}{c|}{}                                                         &                                                     & \multicolumn{1}{c|}{}                                                            &                                                               &                                                          & \multicolumn{1}{c|}{}                                                                    & \small{\attackMitigatedMark} {\tiny \cite{chabanne_protection_2020}} \small{\attackClaimMark} \\ \hline
\small{SMA-*} \tiny{(retraining)} \tiny{\cite{tramer_stealing_2016, reith_efficiently_2019}}                                  &                                                                                                                                   & \multicolumn{1}{c|}{}                                                      & \multicolumn{1}{c|}{}                                                                         & \multicolumn{1}{c|}{}                                                                         & \multicolumn{1}{c|}{}                                                                            & \multicolumn{1}{c|}{}                                                     & \multicolumn{1}{c|}{}                                                             & \multicolumn{1}{c|}{}                                                    & \multicolumn{1}{c|}{\small{\attackMitigatedMark} \tiny{\cite{wang_information_2021}}} &                                                     & \multicolumn{1}{c|}{\small{\attackMitigatedMark} {\tiny \cite{zheng_bdpl_2019}}}         &                                                               &                                                          & \multicolumn{1}{c|}{}                                                                    &                                                                               \\ \hline
\small{SMA-NN} \tiny{\cite{papernot_practical_2017}}                                                                          &                                                                                                                                   & \multicolumn{1}{c|}{}                                                      & \multicolumn{1}{c|}{}                                                                         & \multicolumn{1}{c|}{\small{\attackMitigatedMark} {\tiny \cite{kariyappa_defending_2020}}}             & \multicolumn{1}{c|}{\small{\attackMitigatedMark} {\tiny \cite{orekondy_prediction_2020}}}                & \multicolumn{1}{c|}{\small{\attackMitigatedMark} \tiny{\cite{chen_das-ast_2020}}} & \multicolumn{1}{c|}{\small{\attackLoweredMark} \tiny{\cite{kariyappa_protecting_2021}}}   & \multicolumn{1}{c|}{}                                                    & \multicolumn{1}{c|}{}                                                         &                                                     & \multicolumn{1}{c|}{}                                                            &                                                               &                                                          & \multicolumn{1}{c|}{}                                                                    &                                                                               \\ \hline
\small{SMA-CNN} \tiny{(PD, AL)}  \tiny{\cite{pal_framework_2019}}                                                             &                                                                                                                                   & \multicolumn{1}{c|}{}                                                      & \multicolumn{1}{c|}{\small{\attackLoweredMark} {\tiny \cite{pal_framework_2019}} \small{\speculatemark}}      & \multicolumn{1}{c|}{}                                                                         & \multicolumn{1}{c|}{}                                                                            & \multicolumn{1}{c|}{}                                                     & \multicolumn{1}{c|}{}                                                             & \multicolumn{1}{c|}{}                                                    & \multicolumn{1}{c|}{}                                                         &                                                     & \multicolumn{1}{c|}{}                                                            &                                                               &                                                          & \multicolumn{1}{c|}{\small{\defenceBrokenMark} {\tiny \cite{pal_framework_2019}} \small{\speculatemark}} &                                                                               \\ \hline
\small{SMA-CNN} \tiny{(DS)} \tiny{\cite{mosafi_stealing_2019}}                                                                &                                                                                                                                   & \multicolumn{1}{c|}{}                                                      & \multicolumn{1}{c|}{\small{\defenceBrokenMark} {\tiny \cite{mosafi_stealing_2019}} \small{\attackClaimMark}}  & \multicolumn{1}{c|}{}                                                                         & \multicolumn{1}{c|}{}                                                                            & \multicolumn{1}{c|}{}                                                     & \multicolumn{1}{c|}{}                                                             & \multicolumn{1}{c|}{}                                                    & \multicolumn{1}{c|}{}                                                         &                                                     & \multicolumn{1}{c|}{}                                                            &                                                               &                                                          & \multicolumn{1}{c|}{}                                                                    &                                                                               \\ \hline
\small{SMA-CNN} \tiny{(DS)} \tiny{\cite{yuan_es_2022}}                                                                        & \small{\defenceBrokenMark} {\tiny \cite{yuan_es_2022}}                                                                                    & \multicolumn{1}{c|}{}                                                      & \multicolumn{1}{c|}{}                                                                         & \multicolumn{1}{c|}{}                                                                         & \multicolumn{1}{c|}{}                                                                            & \multicolumn{1}{c|}{}                                                     & \multicolumn{1}{c|}{}                                                             & \multicolumn{1}{c|}{}                                                    & \multicolumn{1}{c|}{}                                                         &                                                     & \multicolumn{1}{c|}{}                                                            &                                                               &                                                          & \multicolumn{1}{c|}{}                                                                    &                                                                               \\ \hline
\small{SMA-CNN;RNN} \tiny{(Activethief)} \tiny{\cite{pal_activethief_2020}}                                                   &                                                                                                                                   & \multicolumn{1}{c|}{}                                                      & \multicolumn{1}{c|}{\small{\defenceBrokenMark} {\tiny \cite{pal_activethief_2020}} \small{\attackClaimMark}}  & \multicolumn{1}{c|}{}                                                                         & \multicolumn{1}{c|}{}                                                                            & \multicolumn{1}{c|}{}                                                     & \multicolumn{1}{c|}{}                                                             & \multicolumn{1}{c|}{}                                                    & \multicolumn{1}{c|}{}                                                         &                                                     & \multicolumn{1}{c|}{}                                                            &                                                               &                                                          & \multicolumn{1}{c|}{}                                                                    &                                                                               \\ \hline
\small{SMA-CNN} \tiny{\cite{juuti_prada_2019}}                                                                                & \small{\defenceBrokenMark} {\tiny \cite{juuti_prada_2019}}                                                                                & \multicolumn{1}{c|}{}                                                      & \multicolumn{1}{c|}{}                                                                         & \multicolumn{1}{c|}{}                                                                         & \multicolumn{1}{c|}{\small{\attackMitigatedMark} {\tiny \cite{orekondy_prediction_2020}}}                & \multicolumn{1}{c|}{\small{\attackMitigatedMark} \tiny{\cite{chen_das-ast_2020}}} & \multicolumn{1}{c|}{\small{\attackLoweredMark} \tiny{\cite{kariyappa_protecting_2021}}}   & \multicolumn{1}{c|}{}                                                    & \multicolumn{1}{c|}{}                                                         &                                                     & \multicolumn{1}{c|}{}                                                            &                                                               &                                                          & \multicolumn{1}{c|}{}                                                                    &                                                                               \\ \hline
\small{SMA-CNN} \tiny{(Knockoff)} \tiny{\cite{orekondy_knockoff_2019}}                                                        & \small{\attackLoweredMark} {\tiny \cite{orekondy_knockoff_2019}}  \small{\attackMitigatedMark} {\tiny \cite{atli_extraction_2020}}                & \multicolumn{1}{c|}{}                                                      & \multicolumn{1}{c|}{}                                                                         & \multicolumn{1}{c|}{\small{\attackMitigatedMark} {\tiny \cite{kariyappa_defending_2020}}}             & \multicolumn{1}{c|}{}                                                                            & \multicolumn{1}{c|}{\small{\attackMitigatedMark} \tiny{\cite{chen_das-ast_2020}}} & \multicolumn{1}{c|}{\small{\attackMitigatedMark} \tiny{\cite{kariyappa_protecting_2021}}} & \multicolumn{1}{c|}{}                                                    & \multicolumn{1}{c|}{}                                                         & \small{\attackMitigatedMark} \tiny{\cite{mazeika_how_2022}} & \multicolumn{1}{c|}{}                                                            &                                                               & \small{\attackMitigatedMark} {\tiny \cite{atli_extraction_2020}} & \multicolumn{1}{c|}{}                                                                    &                                                                               \\ \hline
\small{SMA-CNN} \tiny{\cite{kariyappa_maze_2021}}                                                                             & \small{\attackMitigatedMark} {\tiny \cite{kariyappa_maze_2021}} \small{\attackClaimMark}                                                          & \multicolumn{1}{c|}{}                                                      & \multicolumn{1}{c|}{\small{\attackMitigatedMark} {\tiny \cite{kariyappa_maze_2021}} \small{\attackClaimMark}} & \multicolumn{1}{c|}{\small{\attackMitigatedMark} {\tiny \cite{kariyappa_maze_2021}} \small{\attackClaimMark}} & \multicolumn{1}{c|}{\small{\attackMitigatedMark} {\tiny \cite{kariyappa_maze_2021}} \small{\attackClaimMark}}    & \multicolumn{1}{c|}{}                                                     & \multicolumn{1}{c|}{}                                                             & \multicolumn{1}{c|}{}                                                    & \multicolumn{1}{c|}{}                                                         &                                                     & \multicolumn{1}{c|}{}                                                            &                                                               &                                                          & \multicolumn{1}{c|}{}                                                                    &                                                                               \\ \hline
\small{SMA-RNN} \tiny{\cite{krishna_thieves_2020}}                                                                            & \small{\defenceBrokenMark} {\tiny \cite{krishna_thieves_2020}}                                                                            & \multicolumn{1}{c|}{}                                                      & \multicolumn{1}{c|}{}                                                                         & \multicolumn{1}{c|}{}                                                                         & \multicolumn{1}{c|}{}                                                                            & \multicolumn{1}{c|}{}                                                     & \multicolumn{1}{c|}{}                                                             & \multicolumn{1}{c|}{}                                                    & \multicolumn{1}{c|}{}                                                         &                                                     & \multicolumn{1}{c|}{}                                                            &                                                               &                                                          & \multicolumn{1}{c|}{}                                                                    &                                                                               \\ \hline
\small{SMA-CNN} \tiny{(Black-box Dissector)} \tiny{\cite{wang_black-box_2022}}                                                &                                                                                                                                   & \multicolumn{1}{c|}{}                                                      & \multicolumn{1}{c|}{}                                                                         & \multicolumn{1}{c|}{\small{\defenceBrokenMark} \tiny{\cite{wang_black-box_2022}}}                     & \multicolumn{1}{c|}{\small{\defenceBrokenMark} \tiny{\cite{wang_black-box_2022}}}                        & \multicolumn{1}{c|}{}                                                     & \multicolumn{1}{c|}{}                                                             & \multicolumn{1}{c|}{}                                                    & \multicolumn{1}{c|}{}                                                         &                                                     & \multicolumn{1}{c|}{}                                                            &                                                               &                                                          & \multicolumn{1}{c|}{}                                                                    &                                                                               \\ \hline
\small{SMA-GAN} \tiny{\cite{hu_stealing_2021}}                                                                                &                                                                                                                                   & \multicolumn{1}{c|}{}                                                      & \multicolumn{1}{c|}{}                                                                         & \multicolumn{1}{c|}{}                                                                         & \multicolumn{1}{c|}{}                                                                            & \multicolumn{1}{c|}{}                                                     & \multicolumn{1}{c|}{}                                                             & \multicolumn{1}{c|}{\small{\attackMitigatedMark} \tiny{\cite{hu_stealing_2021}}} & \multicolumn{1}{c|}{}                                                         &                                                     & \multicolumn{1}{c|}{}                                                            &                                                               &                                                          & \multicolumn{1}{c|}{}                                                                    &                                                                               \\ \hline
\small{SMA-Encoder} \tiny{(StolenEncoder)} \tiny{\cite{liu_stolenencoder_2022}}                                               & \small{\defenceBrokenMark} \tiny{\cite{liu_stolenencoder_2022}}                                                                           & \multicolumn{1}{c|}{}                                                      & \multicolumn{1}{c|}{}                                                                         & \multicolumn{1}{c|}{}                                                                         & \multicolumn{1}{c|}{\small{\defenceBrokenMark} \tiny{\cite{liu_stolenencoder_2022}}}                     & \multicolumn{1}{c|}{}                                                     & \multicolumn{1}{c|}{}                                                             & \multicolumn{1}{c|}{}                                                    & \multicolumn{1}{c|}{}                                                         &                                                     & \multicolumn{1}{c|}{}                                                            &                                                               &                                                          & \multicolumn{1}{c|}{}                                                                    &                                                                               \\ \hline
\small{SMA-Encoder} \tiny{\cite{dziedzic_difficulty_2022}}                                                                    &                                                                                                                                   & \multicolumn{1}{c|}{}                                                      & \multicolumn{1}{c|}{}                                                                         & \multicolumn{1}{c|}{}                                                                         & \multicolumn{1}{c|}{\small{\attackLoweredMark} {\tiny \cite{dziedzic_difficulty_2022}} \small{\attackClaimMark}} & \multicolumn{1}{c|}{}                                                     & \multicolumn{1}{c|}{}                                                             & \multicolumn{1}{c|}{}                                                    & \multicolumn{1}{c|}{}                                                         &                                                     & \multicolumn{1}{c|}{}                                                            &                                                               &                                                          & \multicolumn{1}{c|}{}                                                                    &                                                                               \\ \hline
\end{tabular}
}}

\end{table*}

\Cref{tab:comparison-reactive,tab:comparison-proactive} provide a lineup of query-based attacks against reactive and pro-active defences correspondingly, indicating which attack can be mitigated by what defence, and which defence has already been broken by another attack. Rows in \Cref{tab:comparison-reactive,tab:comparison-proactive} correspond to query-based attacks, while columns correspond to defences. Whenever a defence in a column $d$ was shown to mitigate the attack in row $a$, we put a $\attackMitigatedMark$ mark in cell $(a, d)$ and a reference to the respective paper.
If the mitigation was not shown, but a claim has been made without experimental demonstration, we indicate this with an additional \attackClaimMark. If the authors only speculate about the (in)effectiveness of defences, we denote this with an additional \speculatemark.
If a defence was proven or claimed to be completely ineffective against an attack, we indicate this with a $\defenceBrokenMark$ mark. 
A mark \attackLoweredMark means that a defence is partially effective: either the attack performance has fallen only slightly (less than $5\%$), or, if the defence was a monitor, the attack was detected too slowly.
If a defence has been broken at least for one considered dataset, we mark it as broken. 
If a cell $(a, d)$ is empty, it means that there is no information available about the effectiveness of the defence $d$ against the attack $a$, and vice versa.

In this lineup, we did not include papers that present only theoretical results, or do not provide a specific lineup against an attack. For instance, we did not include defence papers which consider their defences against a generic class of attacks without referring to a specific attack paper. 
We also omit defences which are described in attack papers if they are either attack-specific and not effective, or described in insufficient detail. 

From \Cref{tab:comparison-reactive,tab:comparison-proactive}, we can observe that the basic defence method proposed in \cite{tramer_stealing_2016}, which relies on returning labels instead of confidence scores, has been broken by several attacks and, thus, seems unreliable.
Regarding other defences, several early monitors -- while initially successful against early attacks -- have since shown to be ineffective against more recent strategies developed to specifically counter them \cite{kesarwani_model_2018, juuti_prada_2019}. However, none of the monitors published afterwards is broken.
A few defences are, at the time of writing, not broken by specific attacks; however, we note that the current lineup is not complete -- many combinations have simply not been considered, and their outcome would thus be unclear.
Further, we want to mention that there is likely a difference in the exact outcome depending on who is performing the lineup, i.e. whether a novel attack tries to break a defence, or a defence tries to show that it is effective against an attack. Depending on who is in the driver's seat, it might be that the general knowledge of the technique (e.g. the importance of certain parameters) as well as the invested effort might favour the attacker or the defender. 
This again highlights the need for a systematic and "independent" assessment of the effectiveness of defence methods.

\section{Conclusions}
\label{sec:conclusion}
In this paper, we provided a comprehensive overview and systematisation of attacks and defences related to model stealing (model extraction).
We first provide a common terminology, unifying the disparity of notions used across the literature.
We explored the conditions, methodology, and goals of model stealing approaches and subsequently classified them, showing which attacks are possible in specific settings. 
This resulted in a comprehensive taxonomy and guideline.
Moreover, we extensively analysed defence approaches and developed guidelines for choosing the most effective defence strategy. 
We then compared which defence is mitigating -- resp. broken by -- current attack strategies.
Based on our survey and analysis, we observe several research issues, challenges, and future directions of research.

We observe a general \textit{lack of standardised and systematic methodology in the research on model stealing}, and especially on reporting of both attacks and defences. 
For attacks, many works omit important details on the efficiency of the attacks. For example, for query-based attacks, it is often not stated how many queries are required, or how complex the model to be stolen is. Thus, it is difficult to compare efficiency between different methods.

We thus propose a methodology for conducting research in this field, based on the contributions in this paper:
(1) Research papers should use a \textbf{common notation}; the terminology and taxonomies for attacks and defences proposed in this paper would be fitting candidates.
(2) Using this terminology, research contributions should provide a \textbf{detailed threat model}, specifying the goal of the attacker (e.g. following our taxonomy) as well as their capabilities, e.g. the knowledge of the model, the actions (e.g. querying or side-channel), and the resources (e.g. the query budget).
(3) Having defined a concrete goal motivates which criteria should be used to \textbf{measure the attack effectiveness and -efficiency}. 
If no concrete goal can be defined, e.g. to not limit the attack to a specific application, we do encourage future works to be more comprehensive and measure all applicable criteria we have outlined in \Cref{sec:performance_objectives}, such as fidelity, accuracy and transferability, to enable broad comparison with other works.
We note that the methodology in evaluation and reporting results lags behind especially for side-channel attacks, which currently mostly consists of anecdotal evidence and qualitative claims, but does not offer much in terms of quantitative, dependable results.

In terms of methodology, a \textit{unified evaluation framework} would enable more comparability and, thus, progress in techniques; this should include e.g. benchmark datasets, trained models, and open source- and unified implementations of attack- and defence approaches, as for instance in a recent initiative for adversarial examples \cite{croce_robustbench_2021}.

Regarding defences and their performance against various attacks, we observe that the \textit{current experimental lineup between attacks and defences is rather incomplete}. As many combinations were not studied, it is difficult to obtain a clear picture of the overall effectiveness of the proposed methods. A \textbf{large-scale, analytic and empirical evaluation} would help to significantly advance this aspect.
As future research challenge, \textit{adaptive attackers} -- who are widely studied in evasion attacks against Machine Learning models \cite{tramer_adaptive_2020} and constitute attackers who are aware of a certain defence and try to counter it -- are not yet widely considered in model stealing.
For example, against dynamic watermarking, an attacker could be utilising defences against data-poisoning-based backdoor attacks, such as \cite{liu_fine-pruning_2018}.

Another future research challenge is the rather large \textit{gap of dedicated defences against side-channel attacks} (SCA). The main strategy currently used is to rely on classical IT security measures, i.e. a stringent access control, both on the hardware- and software level, or the use of dedicated infrastructure (or effective isolation of processes) to avoid attacks that are exploiting e.g. shared memory access. As these strategies are likely limited in scope, there is, however, a need for pro-active methods that are model-inherent and can reduce the effectiveness of SCAs.

Overall, we note that there are both a large number of possible attacks against the investment made into the creation of trained ML models, but also defence options -- but both are still requiring substantially more structured investigation. 
However, having a structured taxonomy allows to more systematically address the problem space and provides guidance not only for the attacker, but also the defender. 
Failing to consider and invest resources into these challenges may well lead to a shocking awakening if and when some attacker might subtly tell us they know (in much more detail than we wanted them to) what we invested in our GPU cycles in last summer, and that our model was gone in 60(k) queries -- or to allow \textit{us} to tell \textit{them} that \textit{we} know what \textit{they} tried to do to our model last summer.

\begin{acks}
This work received funding from the \grantsponsor{H2020}{European Union's Horizon 2020 research and innovation programme} under grant agreement No \grantnum{H2020}{826078} (FeatureCloud). 
This publication reflects only the authors’ view and the European Commission is not responsible for any use that may be made of the information it contains.
SBA Research (SBA-K1) is a COMET Center within the COMET - Competence Centers for Excellent Technologies Programme and funded by BMK, BMAW, and the federal state of Vienna. COMET is managed by FFG.
\end{acks}

\bibliographystyle{unsrt}
\bibliography{main}

\end{document}